\documentclass{article}
\usepackage[table]{xcolor}
\definecolor{TableOneInk}{HTML}{243B53}
\usepackage{iclr2026_conference,times}

\usepackage{hyperref}
\usepackage{url}

\usepackage{pifont}
\usepackage{tabularx}
\usepackage{multirow}
\usepackage{caption}
\usepackage{booktabs}
\usepackage{graphicx}
\usepackage{makecell}
\usepackage{capt-of}
\usepackage{graphicx}
\usepackage{wrapfig}
\usepackage{amssymb}
\usepackage{hyperref}
\usepackage{booktabs}
\usepackage{graphicx}
\usepackage{amsmath}
\usepackage{algorithm}
\usepackage{algorithmic}
\usepackage{microtype}

\title{What Does an Agent Need? Bundle-Aware Agent Resource Composition}

\iclrfinalcopy
\author{\textbf{Yunxiao Shi}$^{1}$ \quad \textbf{Tingwei Chen}$^{2}$ \quad \textbf{Dannong Xu}$^{1}$ \quad \textbf{Ahad N. Zehmakan}$^{1}$\\
\textbf{Jialie Shen}$^{3}$ \quad \textbf{Zhendong Chu}$^{4}$ \quad \textbf{Qingsong Wen}$^{4}$\\[0.4em]
{\normalfont\small $^{1}$Australian National University \quad $^{2}$University of Technology Sydney}\\
{\normalfont\small $^{3}$City St George's, University of London \quad $^{4}$Squirrel Ai Learning}}
\hypersetup{
  pdftitle={What Does an Agent Need? Bundle-Aware Agent Resource Composition},
  pdfauthor={Yunxiao Shi, Tingwei Chen, Dannong Xu, Ahad N. Zehmakan, Jialie Shen, Zhendong Chu, Qingsong Wen},
  hidelinks
}

\begin{document}

\maketitle

\begin{abstract}
As catalogs of models, tools, and reusable skills expand, determining
which resources an agent should use becomes increasingly
challenging. The value of a resource depends on the backbone and the other
resources it is combined with, and it is known only after execution.
Existing methods address parts of this problem. Token-based tool generators
tie each tool to a learned token, so they select only from the training catalog. Retrievers and routers score each tool or
backbone on its own and miss complementary resources, and bundle reuse and
agent recommendation return configurations built for other tasks. How to
compose heterogeneous resources for a new task remains underexplored.
We formulate \textbf{Task-Conditioned Agent Resource Composition
(AgentComposition)}: jointly selecting a backbone model and composing a
tool--skill bundle that maximize end-to-end execution utility. We propose
\textbf{Retrieval-Grounded Bundle-Aware Agent Generation}. It retrieves
candidate resources and past configurations of similar tasks, and a fine-tuned
generator that reads each resource from its description recombines them by
constrained beam search, so new resources need no retraining. A bundle-aware
critic ranks the complete configurations. It learns from two proxies for real
execution, an LLM teacher and historically successful configurations, and
adapts to end-to-end execution feedback when it is available.
We evaluate by reference recovery, offline execution with judged and native
rewards, and online recommendation with real execution, and we build
AgentSelect-Exec with 5,060 executed configurations for this purpose.
Our method recovers the complete reference configuration within ten candidates
for 67.73\% of AgentSelect queries, nearly twice the strongest baseline, and
leads all baselines on SkillsBench. The proxy-trained critic already raises the
execution score of the top-ranked agent from 7.58 to 8.37 out of 10, and
execution feedback lifts it to 8.78, above the strongest baselines. Four
target-domain queries suffice to adapt the generator to SkillsBench, and agents
recommended online execute better than those of the strongest baselines under
both an LLM judge and native verifiers.
These results establish resource composition as a learnable and transferable
capability, shifting the focus from individual resource relevance to joint task
suitability, toward agents that adapt to changing resource ecosystems and evolve
through accumulated task experience.
\end{abstract}

\section{Introduction}

Large language models (LLMs) have grown from text completion~\cite{Radford2019LanguageMA,brown2020language} into agents that reason, call functions, and act in environments~\cite{yao2023react,schick2023toolformer,qin2024toolllm}. The Model Context Protocol (MCP)~\cite{anthropic2024mcp} and Agent Skills~\cite{anthropic2025agentskills,AgentSkills} expose growing catalogs of tools, data, and reusable skills, and agent harnesses organize them around a backbone model~\cite{MicrosoftAgentHarness,OpenAIAgentHarness,AnthropicAgentHarness}. Choosing these resources well makes agents more adaptive~\cite{toolspectrum,etapp}, keeps the tool context small~\cite{toolscope}, and lets agents keep up with changing ecosystems~\cite{wu-etal-2026-beyond-static}. Each new task therefore raises a practical question: \textbf{which backbone and which resources should an agent use together?}

Existing work answers parts of this question. Tool learning and tool retrieval select tools for a fixed backbone~\cite{ToolLLM,ToolkenGPT,ToolGen,du2024anytool,qu2024completeness,yuan2024craft}, model routing chooses a backbone without tools~\cite{RouteLLM}, and AgentSelect recommends predefined configurations~\cite{AgentSelect}. Composing a configuration for a new task is harder, for four reasons. First, the value of a resource depends on the resources it is combined with. \textit{A seven-day, pet-friendly road trip under a fixed budget needs accommodation search, pet-policy checks, route planning, and cost calculation. Accommodation tools alone leave needs unmet, and a past road-trip configuration may miss the pet constraint.} Selecting resources one by one misses such bundle effects (Appendix~\ref{app:case_study}). Second, the number of configurations grows combinatorially with the catalog. Third, the catalog keeps changing, so methods tied to a fixed set of resources must be retrained. Fourth, a configuration's quality is known only after a costly execution.

We formulate \textbf{Task-Conditioned Agent Resource Composition (AgentComposition)}: given a task and its context, select a backbone and compose a tool--skill bundle that maximize end-to-end execution utility. Agent architecture search such as AgentSquare~\cite{AgentSquare} optimizes how an agent plans, reasons, and remembers. AgentComposition chooses the resources an agent uses. We propose \textbf{Retrieval-Grounded Bundle-Aware Agent Generation}, whose components address these challenges. Retrieval narrows the search to relevant resources and supplies past configurations of similar tasks. A fine-tuned generator composes configurations by constrained beam search, choosing each tool given the backbone and the tools already chosen. It reads each resource from its description, so a new resource is usable as soon as it is indexed. A bundle-level critic estimates execution quality from proxy signals, adapts to execution feedback when available, and reranks the completed configurations.

We evaluate on AgentSelect~\cite{AgentSelect} and SkillsBench~\cite{SkillsBench} and on three datasets with execution outcomes: BFCL~\cite{BFCL}, SWE-Skills-Bench~\cite{SWESkillsBench}, and AgentSelect-Exec, which we build from 5,060 executed configurations. Four findings emerge. (i) Our method recovers complete configurations best, while historical bundles, tool retrievers and model routers, token-based tool generators, bundle generators, and retrieval-augmented LLMs each fall short in ways their designs predict. (ii) Execution feedback makes the critic's rankings agree with execution. The critic then selects better-executing configurations than all proxy-trained variants and the strongest baselines, and it works best as a final reranker. (iii) The learned composition transfers to new domains. Four target queries are enough to adapt the generator to SkillsBench, and Stage~3 adapts the critic when executed candidates are available. (iv) Agents recommended online execute better than those of the strongest baselines, under both an LLM judge and native verifiers, and within our lists, higher-ranked recommendations execute better.

Our contributions are as follows.
\ding{182} We formulate \textbf{AgentComposition}, which composes a backbone and a tool--skill bundle for each task to maximize execution utility, and we identify its bundle-level challenges.
\ding{183} We propose a retrieval-grounded, bundle-aware framework that composes valid configurations from a changing catalog and ranks them with a critic that adapts to execution feedback.
\ding{184} We build AgentSelect-Exec and evaluate by reference recovery, offline execution, transfer to new domains, and online execution\footnote{Code: \url{https://github.com/Ancientshi/AgentComposition}.}.

\section{Related Work}
\label{subsec:related_work}

\noindent \textbf{Agent Construction and Resource Composition.}
Agent construction decides what an agent can do and how it does it.
AgentSquare searches over planning, reasoning, tool-use, and memory modules~\cite{AgentSquare}.
ADAS repeatedly writes and tests agent programs~\cite{hu2025adas}.
SkillWeaver discovers and refines reusable skill APIs~\cite{zheng2025skillweaver},
and PSEC combines skill-specific LoRA modules in parameter space~\cite{PSEC}.
These methods change the agent's design or build new capabilities. We
compose existing catalog resources for a task, choosing a backbone and a
tool--skill bundle together. Model routing chooses between a strong and a weak LLM without tools~\cite{RouteLLM}.

\noindent \textbf{Agent Resource Retrieval and Selection.}
Existing retrieval typically ranks individual resources by their relevance to a query~\cite{ToolRet,efficient_tool_rep}. PORTS aligns a dense tool retriever with LLM preferences over tool selection~\cite{PORTS}, and AnyTool collects APIs for a fixed LLM through hierarchical LLM agents~\cite{du2024anytool}. Scoring tools one at a time finds useful tools but ignores how they work together, so the top-ranked tools form bundles with many extra tools. Generative methods such as ToolkenGPT, ToolGen, and ToolWeaver let a fixed backbone select tools by generating them~\cite{ToolkenGPT,ToolGen,ToolWeaver}. ToolWeaver encodes tool co-usage in hierarchical tool codes that are fixed for the whole catalog. These methods give each tool a learned token or code tied to the training catalog, so new tools need new tokens and retraining. Toolken+ reranks ToolkenGPT's top-$k$ tools by their documentation and adds a reject option that skips the current call~\cite{Toolkenplus}. It improves individual tool calls during generation. Our method treats resource selection as composition. It selects a backbone and a bundle together and generates configurations from retrieved resources and their descriptions, so a new resource can be used as soon as it is indexed.

\noindent \textbf{Bundle Recommendation and Agent Discovery.}
Bundle recommendation ranks existing bundles or constructs new ones~\cite{BundleRecSurvey}.
BGCN models user--item, user--bundle, and bundle--item interactions through
graph convolutions~\cite{BGCN}. CrossCBR aligns item- and bundle-level
representations~\cite{CrossCBR}. Construction methods include
query-conditioned Text2Bundle~\cite{Text2Bundle}, LLM-based adaptive
in-context learning~\cite{AICL}, and discrete-diffusion generation of
unordered bundles~\cite{tu2026discrete}. In agent discovery, AgentSearchBench and AgentSelect retrieve or rank predefined agents or configurations from an existing catalog~\cite{AgentSearchBench,AgentSelect}. We treat agent resources as parts of a bundle and generate a new configuration for each task.

\section{Method}
\label{sec:method}

\noindent\textbf{Problem formulation.}
\emph{AgentComposition} maps a task to an agent that executes it end to end.
A task $Q$ contains a user request and its context, such as relevant session
history. Given backbone models $\mathcal{M}$ and tools and skills
$\mathcal{T}$, an agent configuration $A=(M,B)$ pairs a backbone
$M\in\mathcal{M}$ with a bundle $B\subseteq\mathcal{T}$ of at most $L_{\max}$
tools. Executing $A$ on $Q$ yields a utility $U(Q,A)$, such as a verifier
reward or a judged score. The goal is the configuration with the highest
utility,
\begin{equation*}
A^{\star}(Q)=\operatorname*{arg\,max}_{M\in\mathcal{M},\;B\subseteq\mathcal{T},\;|B|\leq L_{\max}}
U\bigl(Q,(M,B)\bigr).
\end{equation*}
Three properties make this a bundle composition problem.
First, utility is non-additive. The gain of adding a tool $t$,
$U(Q,(M,B\cup\{t\}))-U(Q,(M,B))$, depends on the backbone $M$ and on the
tools already in $B$, which can complement or duplicate $t$
(Appendix~\ref{app:case_study}). Retrieval-based selection scores each
resource on its own and picks the backbone and the bundle in separate steps.
It thus treats this gain as a fixed relevance of $t$ to $Q$, independent of
$M$ and $B$. AgentComposition selects the backbone and the bundle
together, as one configuration.
Second, the search space grows combinatorially with the catalog, to
$|\mathcal{M}|\sum_{m=0}^{L_{\max}}\binom{|\mathcal{T}|}{m}$ configurations.
Third, $U$ is observed only after costly execution. We therefore estimate $U$
from proxy signals, and from execution feedback when it is available. The
catalog also changes over time, so new resources must be usable without
retraining. The output is a list of up to $K$ distinct configurations ranked by estimated
utility, $\mathcal{G}(Q;\mathcal{M},\mathcal{T})=(A_Q^{(1)},\ldots,A_Q^{(K)})$.

\begin{figure*}[h!]
\centering
\vspace{-1mm}
\includegraphics[width=0.9\textwidth]{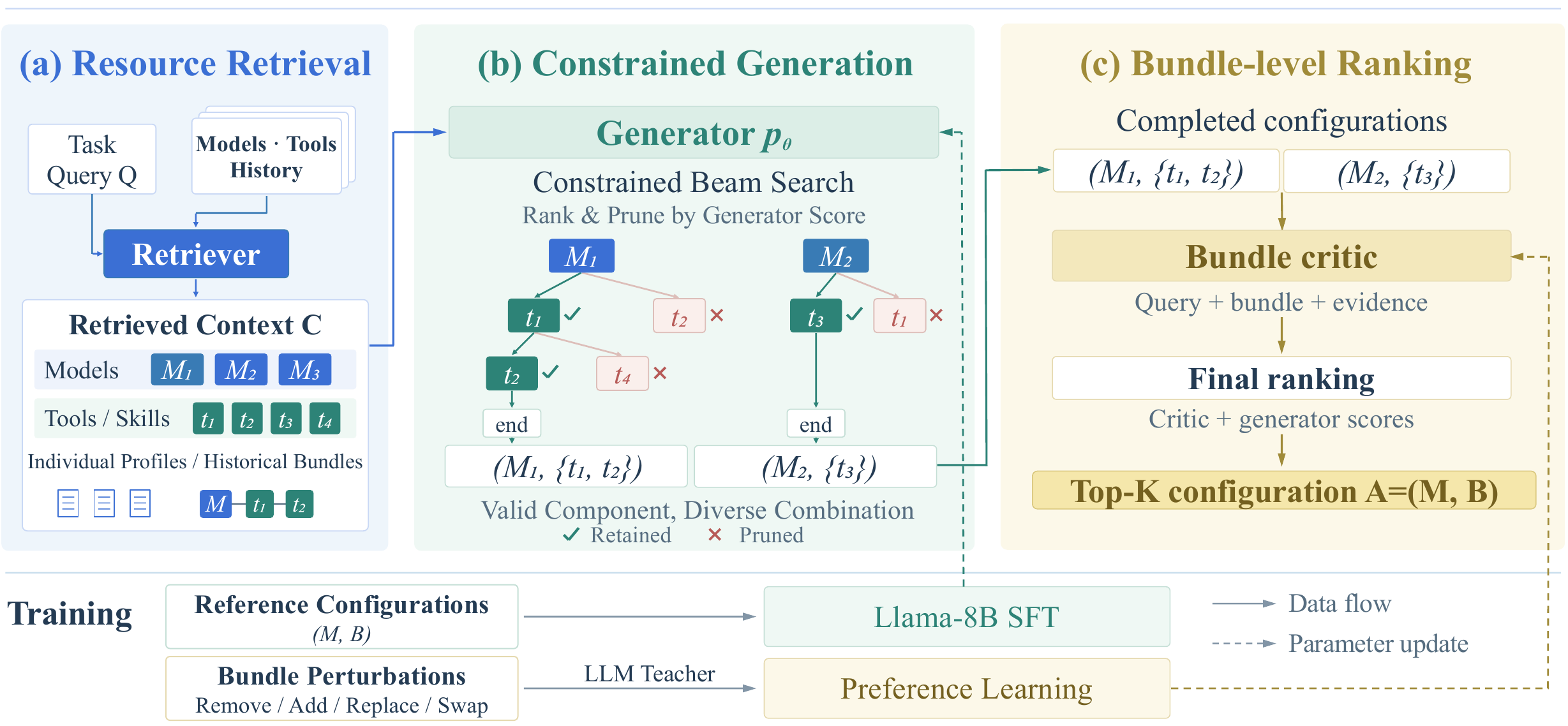}
\vspace{-2mm}
\caption{Overview of our framework: (a) candidate retrieval, (b) constrained search with generator-based pruning, and (c) final reranking. The lower panel shows generator and critic training, with solid arrows for data flow and dashed arrows for parameter updates.}
\label{fig:framework}
\end{figure*}

\noindent\textbf{Method overview.}
Our framework has three steps (Figure~\ref{fig:framework}).
Retrieval finds task-relevant resources and the past configurations of
similar tasks.
A fine-tuned generator then recombines these resources into candidate
configurations through constrained beam search.
Finally, a bundle-aware critic ranks the completed configurations. The critic stands in for execution. It learns from proxy signals and can adapt to execution feedback when it is available. Algorithm~\ref{alg:bundle_aware_search} in
Appendix~\ref{app:bundle_aware_search} gives the procedure.

\subsection{Retrieval Context}
\label{subsec:candidate_retrieval}

Retrieval provides the generator and critic with
$\mathcal{C}(Q)=(\mathcal{M}_r(Q),\mathcal{T}_r(Q),\mathcal{E}_r(Q))$:
retrieved LLMs, retrieved tools, and evidence, which includes resource
profiles, normalized scores, provenance, and complete historical bundles when
available. Semantic retrieval is the default, and a collaborative-signal retriever is added when query--agent
interaction histories exist
(Appendix~\ref{app:retrieval_design}).

\subsection{Structured Generator Fine-Tuning}
\label{subsec:generator_sft}

We initialize $p_{\theta}$ from Meta-Llama-3-8B, apply LoRA to attention and
feed-forward layers, and train the input embeddings and output projection, keeping other pretrained parameters frozen. This adapts the generator to resource
descriptions and the structured output format. Each training instance has a query $Q^{(n)}$, context $\mathcal{C}(Q^{(n)})$,
and reference $A^{(n)\star}=(M^{(n)\star},B^{(n)\star})$. The retrieved context
includes the reference resources. Constrained decoding limits outputs to the
components in this context, so every output uses valid component names. We serialize
$Y(A)=\langle M,\tau_{\mathrm{sep}},\operatorname{Ser}(B),\tau_{\mathrm{end}}\rangle$.
For $B=\{t_1,\ldots,t_m\}\neq\emptyset$,
$\operatorname{Ser}(B)=\langle t_1,\ldots,t_m\rangle$ follows the supervised order. For an allowed empty bundle, it is $\langle\tau_{\emptyset}\rangle$.
The markers $\tau_{\mathrm{sep}}$, $\tau_{\emptyset}$, and
$\tau_{\mathrm{end}}$ separate the backbone from the tools, mark an empty bundle, and end the
configuration. The decoding order affects what each token is conditioned on,
but the bundle itself is an unordered set. Resource names use the existing subword tokens of the vocabulary.

Let $x^{(n)}=\operatorname{Prompt}(I,Q^{(n)},\mathcal{C}(Q^{(n)}))$, where $I$ is a fixed instruction requesting only a configuration. The target $y^{(n)\star}$ is the tokenization of $Y(A^{(n)\star})$ followed by the model's end-of-sequence token. We minimize
\begin{equation}
\mathcal{L}_{\mathrm{SFT}}(\theta)
=-\sum_{n=1}^{N}\sum_{i=1}^{|y^{(n)\star}|}
\log p_{\theta}\!\left(y_i^{(n)\star}\mid x^{(n)},y_{<i}^{(n)\star}\right).
\label{eq:generator_sft}
\end{equation}

\subsection{Bundle-Level Preference Modeling}
\label{subsec:bundle_preference}

The critic $\widehat{U}_{\psi}(Q,A)$ estimates the execution utility
$U(Q,A)$ of a complete configuration. It has an encoder
$\operatorname{Enc}_{\psi_e}$, initialized from AgentEmbedding~\cite{AgentEmbedding}
(an EasyRec model tuned for agent-resource retrieval), and a scalar head
$g_{\psi_s}$. The encoder reads $Q$, $A$, and the retrieved evidence
$\Gamma(A,\mathcal{E}_r(Q))$ for $A$'s resources. We train the critic in
three stages: on LLM-teacher judgments, on counterfactual perturbations judged
against reference configurations, and on execution feedback when it is
available.

\noindent\textbf{Stage 1: Teacher-supervised preferences.}
For each training query, we perturb its reference configuration by removing,
adding, or replacing tools, or by changing the backbone. An LLM teacher scores
each configuration with $r_i\in[0,1]$. We sample same-query pairs $(i,j)$ with
$r_i-r_j\geq0.05$ and cap the number of pairs that change the backbone, the
tools, or both. With $\widehat{U}_i=\widehat{U}_{\psi}(Q_i,A_i)$, we minimize on
minibatch $\mathcal{I}_1$
\begin{equation}
\mathcal{L}_{\mathrm{critic}}^{(1)}
=-\sum_{(i,j)\in\mathcal{I}_1}w_{ij}\log\sigma(\widehat{U}_i-\widehat{U}_j),
\label{eq:critic_rank_loss}
\end{equation}
where $\sigma$ is the logistic sigmoid and the weights
$w_{ij}\propto\sqrt{\max(r_i-r_j,0.05)}$ sum to one. We add 0.1 times the mean
squared error between $\sigma(\widehat{U})$ and the teacher scores.

\noindent\textbf{Stage 2: Reference-based refinement.}
With the backbone fixed, we build preference pairs from tool perturbations that
move a configuration toward a reference configuration: removing an extra tool,
restoring a missing tool, or replacing a tool at a fixed bundle size. A pair is
kept when the perturbation raises precision or recall against that reference
without lowering the other. For backbone changes, we keep the teacher comparisons.
Over a minibatch $\mathcal{I}_2$, we minimize the margin loss
\begin{equation}
\mathcal{L}_{\mathrm{critic}}^{(2)}
=\frac{1}{|\mathcal{I}_2|}\sum_{(i,j)\in\mathcal{I}_2}
\log\!\left(1+\exp\!\left[m-(\widehat{U}_i-\widehat{U}_j)\right]\right),
\qquad m=0.1.
\label{eq:critic_objective}
\end{equation}

\noindent\textbf{Stage 3: Execution-feedback adaptation.}
When execution outcomes are available
(Section~\ref{sec:execution_adaptation}), we fine-tune the Stage~1+2 critic with
\mbox{$\mathcal{L}_{\mathrm{critic}}^{(3)}=\lVert\sigma(\widehat{U})-y\rVert_1+\mathcal{L}_{\mathrm{critic}}^{(2)}$},
where $y\in[0,1]$ is the execution score and Eq.~\eqref{eq:critic_objective}
uses same-query pairs with $y_i>y_j$.

\subsection{Bundle-Aware Constrained Generation}
\label{subsec:bundle_generation}

Search composes a configuration step by step over the retrieved resources. It
selects a backbone first and then adds one tool at a time, so each tool is
chosen given the backbone and the tools already in the bundle.
Two quantities control the search. The proposal budget $P$ is the maximum
number of proposals kept at each expansion, and the pruning width $W(\eta)$ is
the number of paths kept from $\eta$ candidates (Eq.~\eqref{eq:dynamic_width}).

\noindent\textbf{Constrained decoding.}
Each hypothesis $h$ holds a backbone $M_h\in\mathcal{M}_r(Q)$, a prefix
$\rho_h=\langle M_h,\allowbreak\tau_{\mathrm{sep}},\allowbreak t_1,\ldots,t_{\ell}\rangle$,
and a tool set $B_h$. A legal action adds an unused tool from
$\mathcal{T}_r(Q)\setminus B_h$ or terminates the bundle with
$\tau_{\mathrm{end}}$, or with $\tau_{\emptyset},\tau_{\mathrm{end}}$ for an
empty bundle. A path also ends when $|B_h|=L_{\max}$. Prefix tries over the
identifiers restrict decoding to legal actions, so every output is a valid
configuration of retrieved resources.

\noindent\textbf{Action and path scoring.}
Given $x=\operatorname{Prompt}(I,Q,\mathcal{C}(Q))$, an action $a$ with tokens
$\pi(a)=\langle v_1^a,\ldots,v_{|\pi(a)|}^a\rangle$ is scored by its
length-normalized log-likelihood
\begin{equation}
\bar{\ell}_{\theta}(a\mid x,\rho)
=\frac{1}{|\pi(a)|}\sum_{j=1}^{|\pi(a)|}
\log p_{\theta}\!\left(v_j^a\mid x,\rho,v_{<j}^a\right),
\label{eq:phrase_log_probability}
\end{equation}
where $\rho$ is empty for a backbone and $\rho_h$ for later actions.
The empty-bundle action $\tau_{\emptyset}$ is legal only at the first tool
step, where its proposal score is
$\bar{\ell}_{\theta}(\tau_{\emptyset}\mid x,\rho_h)-\lambda_{\emptyset}$ with
$\lambda_{\emptyset}\geq0$.
A path $h$ is scored by the mean log-likelihood over the tokens
$\Omega_h$ of its selected actions,
\begin{equation}
\bar{L}_{\theta}(h)
=\frac{1}{|\Omega_h|}\sum_{i\in\Omega_h}
\log p_{\theta}\!\left(y_i\mid x,y_{<i}\right).
\label{eq:path_generator_score}
\end{equation}

\noindent\textbf{Expansion and pruning.}
A pruning step that receives $\eta$ candidates keeps the top $W(\eta)$, with
\begin{equation}
W(\eta)=\min\!\left\{\eta,\ b_{\max},\ \max\!\left(b_{\min},\lceil r_{\mathrm{keep}}\,\eta\rceil\right)\right\},
\label{eq:dynamic_width}
\end{equation}
where $r_{\mathrm{keep}}\in(0,1]$ is the retain ratio and
$1\leq b_{\min}\leq b_{\max}$ bound the width.
Search starts by scoring every backbone in $\mathcal{M}_r(Q)$ with
Eq.~\eqref{eq:phrase_log_probability} and an empty prefix. The top $P$ become
root proposals, and pruning keeps $W(\eta_0)$ of these $\eta_0$ roots as the
initial beam. Retrieval thus supplies the candidate backbones, and the
generator orders them.
At each depth, every path proposes its top $P$ legal actions by
Eq.~\eqref{eq:phrase_log_probability}, and $\tau_{\mathrm{end}}$ is
added for a nonempty bundle if it is not among them. Completed paths enter a
pool. Unfinished paths with the same backbone and tool set are merged into the
one with the best score, and pruning keeps the top $W(\eta)$ of the $\eta$
merged paths by Eq.~\eqref{eq:path_generator_score}. Search ends when no unfinished path
remains.

\noindent\textbf{Final configuration reranking.}
Completed paths are deduplicated by $(M,B)$ into $\mathcal{F}(Q)$, each with
its best path score $\bar{L}_{\theta}(Q,A)$.
The critic ranks them by
$S_{\mathrm{final}}(Q,A)=\widehat{U}_{\psi}(Q,A)$, and we return the top $K$
configurations as $\mathcal{H}_K(Q)$. Without a critic,
$S_{\mathrm{final}}(Q,A)=\bar{L}_{\theta}(Q,A)-\lambda_{\mathrm{len}}|B|$,
where $\lambda_{\mathrm{len}}\geq0$ favors compact bundles.

\section{Experiments}

\subsection{Datasets}
\label{sec:datasets}

\setlength{\columnsep}{8pt}

\begin{table}[!htbp]
\centering
\caption{Dataset statistics.}
\label{tab:dataset_statistics}
\footnotesize
\setlength{\tabcolsep}{2.5pt}
\begin{tabular}{@{}lrrrrrr@{}}
\toprule
& \multicolumn{3}{c}{Queries} & \multicolumn{3}{c}{Pairs} \\
\cmidrule(lr){2-4}\cmidrule(l){5-7}
Dataset & Train & Val. & Test & Train & Val. & Test \\
\midrule
AgentSelect & 5,022 & 627 & 629 & 12,180 & 1,507 & 1,514 \\
AgentSelect-Exec & 200 & 106 & 200 & 2,000 & 1,060 & 2,000 \\
SkillsBench & 19 & -- & 78 & 109 & -- & 506 \\
BFCL & 161 & 35 & 44 & 2,898 & 630 & 792 \\
SWE-Skills-Bench & 690 & 60 & 230 & 5,520 & 480 & 1,840 \\
\bottomrule
\end{tabular}
\end{table}
\textbf{AgentSelect}~\cite{AgentSelect} pairs queries with agent configurations over 231 LLMs and 12,099 tools. \textbf{SkillsBench}~\cite{SkillsBench} runs agents with and without skills on tasks checked by hidden tests. From its successful runs, we build queries over 19 LLMs, 155 skills, and 20 tools and split them by task. We use it as a new resource domain (Section~\ref{sec:cross_domain_transfer}). Three further datasets record the execution outcome of each configuration. \textbf{AgentSelect-Exec} executes the ten top-ranked configurations of each AgentSelect query and scores each run with an LLM judge. We use it for Stage~3 training and the execution analysis (Section~\ref{sec:execution_adaptation}). \textbf{BFCL}~\cite{BFCL} and \textbf{SWE-Skills-Bench}~\cite{SWESkillsBench} provide native rewards for transfer to new domains (Section~\ref{sec:cross_domain_transfer}). Table~\ref{tab:dataset_statistics} summarizes the splits, and Appendix~\ref{app:dataset_details} details their construction.

\subsection{Implementation Details}
\label{sec:implementation}

The generator is Meta-Llama-3-8B fine-tuned with LoRA on AgentSelect. The critic starts from AgentEmbedding~\cite{AgentEmbedding}, is trained by Stages~1 and~2 on AgentSelect, and is adapted by Stage~3 on the AgentSelect-Exec training split. Search uses adaptive beam bounds $(b_{\min},b_{\max})=(1,2)$, retain ratio $r_{\mathrm{keep}}=0.4$, proposal budget $P=5$, $L_{\max}=6$, and $K=10$. The critic reranks completed configurations without guiding search. Appendix~\ref{app:implementation_details} gives all training and search settings, and Appendix~\ref{app:beam_critic_sensitivity} reports the parameter grid.

\subsection{Baselines}
\label{sec:baselines}

We compare five baseline families (Appendix~\ref{app:baselines}).
(i)~\textbf{Retrieval-based composition} pairs a retrieved LLM with independently retrieved tools (Component-Ret) or with a historical bundle (Bundle-Ret).
(ii)~\textbf{Tool retrieval and model routing} covers PORTS~\cite{PORTS} (a tool retriever aligned with LLM preferences), AnyTool~\cite{du2024anytool} (API retrieval by hierarchical LLM agents), and RouteLLM~\cite{RouteLLM} (routing queries between LLMs).
(iii)~\textbf{Token-based tool generation} covers ToolGen~\cite{ToolGen} and ToolWeaver~\cite{ToolWeaver}, which generate learned tool tokens or codes.
(iv)~\textbf{Bundle generation} adapts sequential Text2Bundle~\cite{Text2Bundle} and discrete-diffusion DDBC-Adapt~\cite{tu2026discrete}.
(v)~\textbf{Retrieval-augmented LLMs} generate configurations from the same retrieved evidence as ours, using RAG-Llama, GPT-5.6-Sol/Terra/Luna, DeepSeek-V4-Pro, Qwen3.8-27B, and Kimi-K2.6.
Families (ii) and (iii) select either tools or a backbone, so we complete each configuration with the other module of Component-Ret (Appendix~\ref{app:adapted_baselines}).

\subsection{Evaluation Metrics}
\label{sec:main_metrics}
Each method returns 10 configurations, and metrics are averaged over queries on a 0--100 scale. Our primary metric, \textbf{Rank-Discounted Component Recall (RDCR@10)}, discounts component recall by rank, rewarding partial recovery near the top. \textbf{RDCP@10} similarly discounts component precision and penalizes components absent from the matched reference. \textbf{ToolR@1} is the share of required tools covered at rank 1, and \textbf{CompR@1} also counts the reference LLM. \textbf{Tool-Hit@10} asks whether any candidate covers all required tools. \textbf{CR-Hit@1}, \textbf{CR-Hit@10}, and \textbf{CR-MRR@10} additionally require the reference LLM, with MRR rewarding earlier first hits. Appendix~\ref{app:evaluation_metrics} defines all metrics, and Table~\ref{tab:main_results} reports them.

\begin{table*}[h!]
\centering
\caption{Agent recommendation performance on AgentSelect (\%, higher is better). The top three scores in each column are shaded.}

\label{tab:main_results}

\begingroup

\definecolor{TableOneBlue}{HTML}{3B6FD4}
\definecolor{TableOneBand}{HTML}{F1F4F7}
\definecolor{TableOneInk}{HTML}{243B53}

\newcommand{\tabfirst}[1]{%
  \cellcolor{TableOneBlue!32}\textbf{#1}}
\newcommand{\tabsecond}[1]{%
  \cellcolor{TableOneBlue!16}#1}
\newcommand{\tabthird}[1]{%
  \cellcolor{TableOneBlue!7}#1}

\small
\setlength{\tabcolsep}{2.5pt}
\renewcommand{\arraystretch}{0.98}
\arrayrulecolor{TableOneInk!65}

\begin{tabular}{@{}l*{8}{c}@{}}
\toprule
& \multicolumn{2}{c}{Tool Recovery}
& \multicolumn{4}{c}{Agent Recovery}
& \multicolumn{2}{c}{Ranked Quality} \\

\cmidrule(lr){2-3}
\cmidrule(lr){4-7}
\cmidrule(l){8-9}

\textbf{Method}
& \shortstack{ToolR\\@1}
& \shortstack{Tool-Hit\\@10}
& \shortstack{CompR\\@1}
& \shortstack{CR-Hit\\@1}
& \shortstack{CR-Hit\\@10}
& \shortstack{CR-MRR\\@10}
& \textbf{\shortstack{RDCR\\@10}}
& \textbf{\shortstack{RDCP\\@10}} \\

\midrule

Component-Ret
& 27.15
& 12.40
& 22.61
& 1.91
& 6.04
& 2.92
& 20.72
& 11.78 \\

Bundle-Ret
& \tabsecond{90.06}
& \tabfirst{97.62}
& \tabsecond{66.74}
& \tabsecond{10.33}
& 12.40
& 11.19
& 33.81
& 32.55 \\
\arrayrulecolor{TableOneBlue!25}
\specialrule{0.35pt}{1pt}{1pt}
\arrayrulecolor{TableOneInk!65}

PORTS
& 72.23
& 55.95
& 54.20
& 9.52
& 32.74
& 15.73
& 52.30
& 26.01 \\

AnyTool
& 11.54
& 5.41
& 13.33
& 0.00
& 2.60
& 1.06
& 10.96
& 12.14 \\

RouteLLM
& 27.16
& 14.88
& 24.74
& 5.54
& 9.10
& 6.50
& 21.96
& 10.83 \\
\arrayrulecolor{TableOneBlue!25}
\specialrule{0.35pt}{1pt}{1pt}
\arrayrulecolor{TableOneInk!65}

ToolGen
& 37.95
& 28.40
& 31.91
& 4.16
& 15.60
& 7.10
& 30.28
& 31.05 \\

ToolWeaver
& 21.68
& 14.29
& 19.14
& 0.60
& 5.36
& 1.84
& 18.77
& 19.95 \\

\arrayrulecolor{TableOneBlue!25}
\specialrule{0.35pt}{1pt}{1pt}
\arrayrulecolor{TableOneInk!65}

Text2Bundle
& 40.28
& 31.32
& 31.99
& 2.07
& 3.82
& 2.44
& 31.28
& 17.66 \\

DDBC-Adapt
& 70.61
& 60.25
& 54.50
& 8.90
& 32.43
& 15.61
& 49.92
& 44.73 \\

\arrayrulecolor{TableOneBlue!25}
\specialrule{0.35pt}{1pt}{1pt}
\arrayrulecolor{TableOneInk!65}

RAG-Llama
& 52.68
& 49.92
& 40.58
& 3.97
& 6.04
& 4.48
& 15.79
& 16.23 \\

RAG-GPT-5.6-Sol
& 78.98
& 69.79
& 57.86
& 4.93
& 30.37
& 13.17
& 54.76
& 52.69 \\

RAG-GPT-5.6-Terra
& 81.92
& 71.86
& 60.23
& 7.00
& 31.80
& 14.78
& 56.37
& \tabthird{54.24} \\

RAG-GPT-5.6-Luna
& 74.23
& 62.48
& 55.57
& 6.36
& 26.07
& 12.53
& 51.68
& 50.63 \\

RAG-DeepSeek-V4-Pro
& 73.04
& 58.66
& 54.42
& 6.36
& 27.03
& 12.29
& 52.36
& 43.61 \\

RAG-Qwen3.8-27B
& 82.81
& 73.29
& 61.71
& 7.79
& 31.00
& 14.46
& 57.94
& 40.48 \\

RAG-Kimi-K2.6
& 87.63
& 81.08
& 65.03
& 9.38
& \tabthird{36.73}
& \tabthird{17.10}
& \tabthird{62.83}
& 37.29 \\

\arrayrulecolor{TableOneBlue!25}
\specialrule{0.35pt}{1pt}{1pt}
\arrayrulecolor{TableOneInk!65}

\textbf{Ours} (generator only)
& \tabthird{89.99}
& \tabthird{83.15}
& \tabfirst{66.94}
& \tabfirst{12.24}
& \tabfirst{67.73}
& \tabfirst{29.02}
& \tabfirst{65.88}
& \tabfirst{65.28} \\

\textbf{Ours} (with critic)
& \tabfirst{90.19}
& \tabsecond{87.92}
& \tabthird{66.05}
& \tabthird{9.70}
& \tabsecond{58.51}
& \tabsecond{23.35}
& \tabsecond{65.48}
& \tabsecond{62.71} \\

\bottomrule
\end{tabular}

\endgroup
\vspace{-2mm}
\end{table*}

\subsection{Overall Recommendation Performance}
\label{sec:main_results}

This section tests whether composing a configuration as a whole recovers
complete configurations better than reusing, retrieving, or generating
resources one at a time. Table~\ref{tab:main_results} shows that it does. The
two variants of ours rank first and second in RDCR@10 and RDCP@10, and each
baseline family falls short in ways its design predicts.

\noindent\textbf{Historical bundles hold the right resources, but not the right configuration.}
Bundle-Ret reaches the highest Tool-Hit@10 (97.62\%), so the bundles of
similar past queries contain most required tools. Reused as they are, they
rarely match the whole configuration of a new task (12.40\% CR-Hit@10), so we
recompose the retrieved resources.

\noindent\textbf{Selecting resources one by one does not compose them.}
Retrievers and routers score each component on its own. PORTS finds relevant
tools but adds many that the task does not need (26.01\% RDCP@10), and
token-based generators such as ToolGen and ToolWeaver, which tie each tool to a
learned token, recover even less. RAG-LLMs read the same evidence as ours and
select relevant resources, with high ToolR@1 and CompR@1, but they rarely
recover the backbone and all tools together (at most 36.73\% CR-Hit@10).

\noindent\textbf{Bundle generators fix the backbone or the vocabulary.}
Text2Bundle builds a bundle for a fixed CF-selected LLM and rarely recovers the
whole configuration (3.82\% CR-Hit@10). DDBC-Adapt generates the backbone and
the tools together from the components of its training vocabulary and reaches
32.43\%, about half of ours.

\noindent\textbf{Joint composition recovers complete configurations.}
Our generator chooses each tool given the backbone and the tools already
chosen. It recovers the complete configuration within ten candidates on
67.73\% of queries, nearly twice the best baseline. Constrained decoding
keeps every output valid, whereas 3\% of free outputs name unknown
components (Appendix~\ref{app:decoding_validity}).

\noindent\textbf{Reference recovery does not measure execution.}
Critic reranking slightly lowers recovery, e.g., RDCP@10 from 65.28\% to
62.71\%. The critic estimates execution quality, and a configuration that
departs from the reference can still execute well.
Section~\ref{sec:structural_ablation} therefore evaluates rankings by
execution.

\section{Further Analysis}

\subsection[From Reference Configurations to Execution Signals]{From Reference Configurations to Execution Signals}
\label{sec:structural_ablation}
\label{sec:execution_adaptation}

Our critic is meant to estimate execution quality, which reference recovery
does not capture. On AgentSelect-Exec, we ask which supervision lets the critic
predict execution quality, where in the pipeline it should act, and whether its
full ranking agrees with execution. Execution feedback (Stage~3) lets the critic
pick the best-executing configurations and order the rest as execution does,
and the critic works best as a final reranker.

\begin{figure}[t]
\centering
\includegraphics[width=0.95\linewidth]{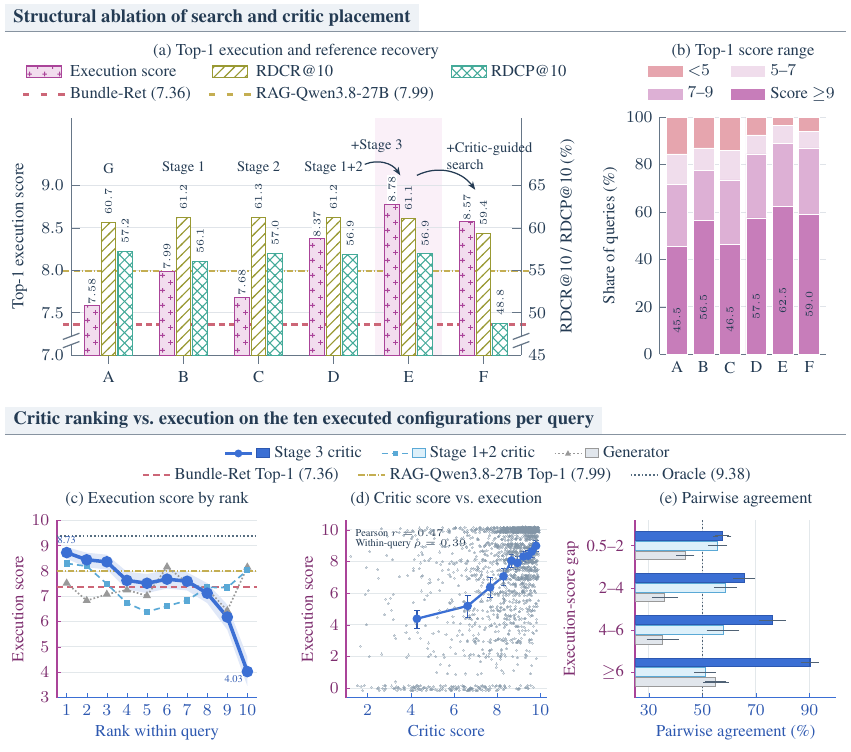}
\caption{Execution analysis on the 200 AgentSelect-Exec test queries. Top-1 scores in (a)--(b) come from re-executing each top-ranked configuration, and (c)--(e) use the recorded executions of AgentSelect-Exec. In (a), purple bars give execution scores, and hatched bars give reference recovery. Bands and error bars are 95\% query-bootstrap intervals.}
\label{fig:structural_ablation}
\end{figure}

Figure~\ref{fig:structural_ablation}a--b compares six rankers on the 200
test queries. A ranks candidates by generator scores (G). B, C, and D rerank them
with critics trained on Stage~1, Stage~2, and both stages. E adds Stage~3, and F
also applies the Stage~3 critic during beam search. A--E rerank the same ten
executed configurations per query (Appendix~\ref{app:agentselect_exec}), and F
ranks the pool of its own search.

\noindent\textbf{Each signal brings the critic closer to execution.}
Both proxy signals improve the top choice over generator scores, and together
they help more (D, 8.37 versus 7.58 for A). Execution feedback helps most.
E reaches 8.78 and cuts the share of queries scoring below 5 from 15.5\%
to 3.5\%. Reference recovery stays flat across A--E (60.7\%--61.3\%
RDCR@10), so it misses these gains. Even the proxy-trained critic D selects
better configurations than Bundle-Ret (7.36) and RAG-Qwen3.8-27B (7.99).

\noindent\textbf{The critic belongs at the end of search.}
Guiding beam search with the Stage~3 critic (F) lowers the top choice to
8.57. The critic is trained on complete configurations, so it misjudges
partial ones and prunes good branches early. We therefore apply it only after search.

\noindent\textbf{The adapted critic orders all candidates as execution does.}
Beyond the top choice (Figure~\ref{fig:structural_ablation}c--e), execution
quality falls steadily with the Stage~3 critic's rank, from 8.73 at rank~1
to 4.03 at rank~10, so configurations that fail in execution sink to the
bottom. The Stage~1+2 critic and the generator spread such failures across their
lists. Within queries, the Stage~3 critic orders 68.0\% of configuration
pairs correctly, against 55.8\% for the Stage~1+2 critic and 43.1\% for
the generator. On pairs whose execution scores differ by at least 6 points,
it is correct 90.2\% of the time (Figure~\ref{fig:structural_ablation}e).

\vspace{-2mm}

\subsection[Generalization to New Domains]{Generalization to New Domains}
\label{sec:cross_domain_transfer}

Our generator reads resources from their descriptions, so its learned
composition should carry over to new domains. We test this on SkillsBench,
BFCL, and SWE-Skills-Bench and ask which component to adapt. The learned
composition transfers. A few target queries are enough to
adapt the generator, and executed candidates are enough to adapt the critic.

\begin{wrapfigure}{r}{0.36\linewidth}
\vspace{-4mm}
\centering
\captionsetup{font=footnotesize,skip=3pt}
\includegraphics[width=\linewidth]{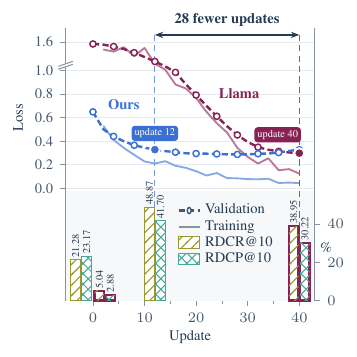}
\caption{Few-shot adaptation on SkillsBench: training and validation loss (lines) and test RDCR@10 and RDCP@10 (bars). Llama is shown in orange.}\label{fig:skillsbench_transfer_loss}
\vspace{-3mm}
\end{wrapfigure}
\noindent On SkillsBench, we fine-tune the AgentSelect-trained generator on the 19
training queries and rank candidates by its scores. A few-shot
variant trains on 4 of them and validates on the other 15. We compare it
with fine-tuning the base Meta-Llama-3-8B
(Figure~\ref{fig:skillsbench_transfer_loss}). BFCL and SWE-Skills-Bench
provide executed configurations with native rewards. There we rank candidates
with the generator (Generator Rank) or the critic (Critic Rank), each with and
without fine-tuning on target training tasks. Stage~3 fine-tunes the
critic. RAG-Qwen3.8-27B ranks the same candidates from task and
resource descriptions.

\noindent\textbf{Composition transfers to a new resource domain.}
On SkillsBench, ours leads all baselines on every metric
(Table~\ref{tab:skillsbench_results}) and recovers the complete configuration
within ten candidates on 47.44\% of the queries, against at most 33.33\%
for the baselines. RAG-LLMs transfer unevenly. Kimi-K2.6, strong on
AgentSelect, falls from 36.73\% to 3.85\% CR-Hit@10. Constrained decoding
matters more in the new domain. Without target training, half of the free
outputs (51.3\%) are invalid.

\begin{table}[H]
\centering
\caption{Agent recommendation performance on SkillsBench (\%, higher is better). The top three scores in each column are shaded.}

\label{tab:skillsbench_results}

\begingroup

\definecolor{TableOneBlue}{HTML}{3B6FD4}
\definecolor{TableOneBand}{HTML}{F1F4F7}
\definecolor{TableOneInk}{HTML}{243B53}

\newcommand{\tabfirst}[1]{%
  \cellcolor{TableOneBlue!32}\textbf{#1}}
\newcommand{\tabsecond}[1]{%
  \cellcolor{TableOneBlue!16}#1}
\newcommand{\tabthird}[1]{%
  \cellcolor{TableOneBlue!7}#1}

\small
\setlength{\tabcolsep}{2.5pt}
\renewcommand{\arraystretch}{0.98}
\arrayrulecolor{TableOneInk!65}

\begin{tabular}{@{}l*{8}{c}@{}}
\toprule
& \multicolumn{2}{c}{Tool Recovery}
& \multicolumn{4}{c}{Agent Recovery}
& \multicolumn{2}{c}{Ranked Quality} \\

\cmidrule(lr){2-3}
\cmidrule(lr){4-7}
\cmidrule(l){8-9}

\textbf{Method}
& \shortstack{ToolR\\@1}
& \shortstack{Tool-Hit\\@10}
& \shortstack{CompR\\@1}
& \shortstack{CR-Hit\\@1}
& \shortstack{CR-Hit\\@10}
& \shortstack{CR-MRR\\@10}
& \textbf{\shortstack{RDCR\\@10}}
& \textbf{\shortstack{RDCP\\@10}} \\

\midrule
RAG-GPT-5.6-Sol
& \tabsecond{71.72}
& \tabsecond{41.03}
& \tabsecond{56.22}
& 0.00
& \tabthird{24.36}
& 4.24
& \tabthird{57.31}
& \tabsecond{54.47} \\

RAG-GPT-5.6-Terra
& 69.71
& \tabthird{32.05}
& 52.62
& \tabthird{1.28}
& 16.67
& \tabthird{4.91}
& 53.86
& 44.62 \\

RAG-GPT-5.6-Luna
& 35.67
& 3.85
& 31.97
& 0.00
& 2.56
& 0.40
& 32.74
& 47.98 \\

RAG-DeepSeek-V4-Pro
& 37.05
& 1.28
& 33.07
& 0.00
& 1.28
& 0.43
& 33.72
& 34.73 \\

RAG-Qwen3.8-27B
& \tabthird{71.02}
& \tabsecond{41.03}
& \tabthird{56.17}
& \tabthird{1.28}
& \tabsecond{33.33}
& \tabsecond{9.73}
& \tabsecond{57.47}
& \tabthird{50.25} \\

RAG-Kimi-K2.6
& 41.76
& 3.85
& 40.14
& \tabsecond{2.56}
& 3.85
& 3.21
& 38.64
& 34.98 \\

\arrayrulecolor{TableOneBlue!25}
\specialrule{0.35pt}{1pt}{1pt}
\arrayrulecolor{TableOneInk!65}

\textbf{Ours}
& \tabfirst{73.61}
& \tabfirst{53.85}
& \tabfirst{61.01}
& \tabfirst{15.38}
& \tabfirst{47.44}
& \tabfirst{24.91}
& \tabfirst{60.31}
& \tabfirst{60.87} \\

\bottomrule
\end{tabular}

\endgroup
\vspace{-1.5mm}
\end{table}

\noindent\textbf{A few queries suffice to adapt the generator.}
Trained on AgentSelect, our generator starts far ahead of the original Llama
(21.28 versus 5.04 RDCR@10) and needs 12 updates instead of 40 to
reach a similar validation loss (Figure~\ref{fig:skillsbench_transfer_loss}). At these
checkpoints it also recovers more of the references (48.87 versus 38.95
RDCR@10). The generator thus carries
the task format and the composition it learned on AgentSelect into the new
domain.

\noindent\textbf{With executed candidates, adapting the critic suffices.}
Without target training, the Stage~1+2 critic already beats RAG-Qwen3.8-27B on
BFCL (90.91 versus 88.64 Top-1) but trails it slightly on
SWE-Skills-Bench (Table~\ref{tab:transfer_execution}). Stage~3 on target
executions lifts the critic to 93.18 on BFCL and 77.13 on
SWE-Skills-Bench, above RAG-Qwen3.8-27B. This matches or beats fine-tuning
the generator.

\begin{table}[H]
\centering
\caption{Offline ranking on BFCL and SWE-Skills-Bench (\%, higher is better). Top-1 is the native reward of the top-ranked configuration. The best score in each column is shaded.}
\label{tab:transfer_execution}
\definecolor{TableOneBlue}{HTML}{3B6FD4}\definecolor{TableOneInk}{HTML}{243B53}
\newcommand{\tabbest}[1]{\cellcolor{TableOneBlue!32}\textbf{#1}}
\small
\setlength{\tabcolsep}{4pt}\arrayrulecolor{TableOneInk!65}%
\begin{tabularx}{\linewidth}{@{}l*{6}{>{\centering\arraybackslash}X}@{}}
\toprule
& \multicolumn{3}{c}{BFCL} & \multicolumn{3}{c}{SWE-Skills-Bench} \\
\cmidrule(lr){2-4}\cmidrule(l){5-7}
\textbf{Method} & Top-1 & RDCR & RDCP & Top-1 & RDCR & RDCP \\
\midrule
RAG-Qwen3.8-27B & 88.64 & 92.75 & 92.75 & 76.85 & \tabbest{71.70} & \tabbest{71.00} \\
\addlinespace[1.5pt]
\multicolumn{7}{@{}l}{\textcolor{TableOneInk!75}{\textit{Ours, without target training}}} \\
\quad Generator Rank & 72.73 & 90.04 & 90.04 & 75.01 & 71.43 & 70.71 \\
\quad Critic Rank & 90.91 & 93.56 & 93.56 & 73.79 & 70.21 & 70.34 \\
\addlinespace[1.5pt]
\multicolumn{7}{@{}l}{\textcolor{TableOneInk!75}{\textit{Ours, with target fine-tuning}}} \\
\quad Generator Rank & 90.91 & 95.68 & 95.68 & \tabbest{77.24} & 70.68 & 70.48 \\
\quad Critic Rank & \tabbest{93.18} & \tabbest{97.02} & \tabbest{97.02} & 77.13 & 70.10 & 70.73 \\
\bottomrule
\end{tabularx}
\vspace{-1.5mm}
\end{table}

\begin{figure}[H]
\vspace{-1.5mm}
\centering
\includegraphics[width=0.95\linewidth]{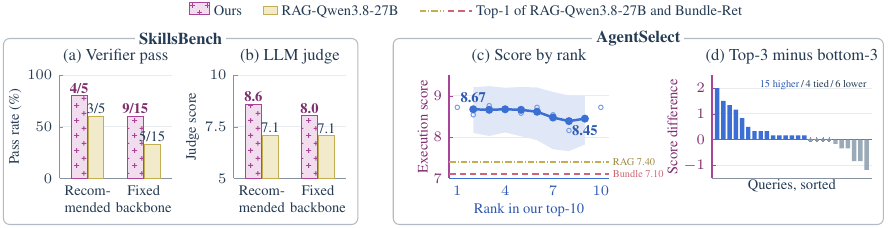}
\caption{Real execution. SkillsBench: (a) verifier pass rate and (b) judge score of the top-ranked agents. AgentSelect: (c) score by rank in our top-10 lists, with baseline top-1 scores as lines, and (d) per query, the mean score of ranks 1--3 minus that of ranks 8--10.}
\label{fig:execution_baselines}
\label{fig:online_execution}
\vspace{-1.5mm}
\end{figure}

\subsection[Online Recommendation and Execution]{Online Recommendation and Execution}
\label{sec:online_execution}

Beyond reranking recorded executions, we ask whether the agents our pipeline
recommends online also execute better. We run
the full pipeline on held-out queries and score each executed recommendation
with an LLM judge and, on SkillsBench, with native verifiers
(Appendix~\ref{app:e2e_execution}). On
SkillsBench, our recommendations execute better than those of RAG-Qwen3.8-27B
under both the native verifier and the LLM judge
(Figure~\ref{fig:online_execution}a--b). They also pass more verifiers when
every method runs on the same fixed backbone, so the gain comes from the
selected skills. On AgentSelect, every rank of our top-10 lists executes better
than the top-1 agents of both baselines (Figure~\ref{fig:online_execution}c).
Within the lists, higher ranks execute better. The score falls from 8.67 at
ranks 1--3 to 8.45 at ranks 8--10, and ranks 1--3 outscore ranks 8--10 on
15 of 25 queries and trail on 6 (Figure~\ref{fig:online_execution}c--d).
Appendix~\ref{app:e2e_native} also ranks official SkillsBench runs offline.

\vspace{-2mm}

\section{Conclusion and Future Work}

We formulate AgentComposition, which composes a backbone and a tool--skill bundle for each task to maximize execution utility, and propose a retrieval-grounded, bundle-aware framework for it. Across reference recovery, offline execution, new domains, and online execution, composing a configuration as a whole outperforms selecting resources one by one, and a critic adapted to execution feedback ranks configurations as execution does. Because the generator reads resources from their descriptions, the learned composition transfers to new domains with a few target queries. Our method is limited by retrieval recall, and its critic relies on proxy signals when no execution feedback is available. Future work will use the adapted critic as a reinforcement-learning reward for the generator.

\clearpage

\bibliography{iclr2026_conference}
\bibliographystyle{iclr2026_conference}

\appendix

\section{AI Usage Statement}
In this work, we used generative AI tools to assist with translation,
dataset cleaning and reformatting. Additionally, we used
these tools to improve the readability of the manuscript, summarize and
analyze existing literature, and identify relevant references. We reviewed
all AI-assisted work, including checking the references suggested by
these tools. We take responsibility for the final content of this work,
including all text, claims, and artifacts produced with the assistance
of generative AI.

\section{Retrieval Design}
\label{app:retrieval_design}

The AgentSelect configuration in Appendix~\ref{app:implementation_details} combines
CF LLM retrieval, CF historical-bundle retrieval, and semantic individual-tool
retrieval. SkillsBench uses the fixed
inventory described in Appendix~\ref{app:implementation_details}.

\noindent\textbf{CF retrieval.}
Let $d_M$ and $d_G$ be profiles of backbone $M\in\mathcal{M}$ and historical
tool bundle $G\in\mathcal{G}\subseteq2^{\mathcal{T}}$. Two trained two-tower
retrievers use TF--IDF features and query-conditioned supervision from
AgentSelect~\cite{AgentSelect}:
\begin{align*}
s_{\mathrm{LLM}}^{\mathrm{sp}}(Q,M)
&=\left\langle g_{\phi_Q}^{\mathrm{LLM}}(\mathrm{TFIDF}(Q)),
 g_{\phi_M}^{\mathrm{LLM}}(\mathrm{TFIDF}(d_M))\right\rangle,\\
s_{\mathrm{bun}}^{\mathrm{sp}}(Q,G)
&=\left\langle g_{\phi_Q}^{\mathrm{bun}}(\mathrm{TFIDF}(Q)),
 g_{\phi_G}^{\mathrm{bun}}(\mathrm{TFIDF}(d_G))\right\rangle.
\end{align*}
Historical bundles remain intact as retrieval evidence; their tools are also
merged and deduplicated into the candidate tool inventory.

\noindent\textbf{Semantic retrieval.}
Dense encoders match query text to component profiles. We use tuned
EasyRec models~\cite{easyrec} trained with AgentSelect supervision; the
representations are further studied in AgentEmbedding~\cite{AgentEmbedding}.
For $u=M$ or $t$, with corresponding component encoder $e_u$, the score is
$\langle e_Q(Q),e_u(d_u)\rangle$.

\noindent\textbf{Context assembly.}
The selected rankings provide $\mathcal{M}_r(Q)$ and $\mathcal{T}_r(Q)$.
Search roots drawn from $\mathcal{M}_r(Q)$ are ordered by the generator
score (Section~\ref{subsec:bundle_generation}).
Evidence $\mathcal{E}_r(Q)$ keeps component profiles,
normalized scores, source branches, and retrieved historical bundles.
On AgentSelect, the three retrieval outputs occupy separate context
sections, with shortened profiles to fit the input budget.
Appendix~\ref{app:retrieval_analysis} measures the coverage of each branch.

\section{Constrained Generation and Bundle-Critic Reranking}
\label{app:bundle_aware_search}

Algorithm~\ref{alg:bundle_aware_search} gives the search and final ranking of
Section~\ref{subsec:bundle_generation}. Generator scores guide expansion and
pruning, and the critic ranks the completed configurations.
Every expansion, including the root step, keeps at most $P$ proposals scored
by Eq.~\eqref{eq:phrase_log_probability}, and every pruning step keeps
$W(\eta)$ of its $\eta$ inputs (Eq.~\eqref{eq:dynamic_width}).

\begin{algorithm}[h]
\caption{Constrained Generation with Optional Bundle-Critic Reranking}
\label{alg:bundle_aware_search}
\begin{algorithmic}[1]
\REQUIRE Query $Q$, context $\mathcal{C}(Q)$, generator $p_{\theta}$,
optional critic $\widehat{U}_{\psi}$,
proposal budget $P\geq1$, pruning width $W(\cdot)$
(Eq.~\eqref{eq:dynamic_width}),
tool limit $L_{\max}$, output size $K\geq1$
\ENSURE Up to $K$ distinct ranked configurations $\mathcal{H}_K(Q)$
\STATE Score each backbone in $\mathcal{M}_r(Q)$ by Eq.~\eqref{eq:phrase_log_probability},
keep the top $P$ as root proposals, and keep the top $W(\eta_0)$ of these
$\eta_0$ roots, each followed by $\tau_{\mathrm{sep}}$, as the initial beam
\STATE $\mathcal{F}_{\mathrm{paths}}\leftarrow\emptyset$ \hfill(completed pool)
\STATE If $L_{\max}=0$, complete the roots with the empty-bundle
representation and move them to $\mathcal{F}_{\mathrm{paths}}$
\WHILE{the beam is nonempty}
    \STATE Enumerate the legal unused-tool and termination actions of each path
    \STATE Keep the top $P$ legal actions of each path by
    Eq.~\eqref{eq:phrase_log_probability}, with the no-tool penalty
    $\lambda_{\emptyset}$ on $\tau_{\emptyset}$, and add
    $\tau_{\mathrm{end}}$ of a nonempty bundle if it is not among them
    \STATE Extend the paths and update their scores by Eq.~\eqref{eq:path_generator_score}
    \STATE Move terminated paths and paths with $L_{\max}$ tools to $\mathcal{F}_{\mathrm{paths}}$
    \STATE Merge unfinished paths with the same $(M,B)$ into the best ordering, and
    keep the top $W(\eta)$ of the $\eta$ merged paths by $\bar{L}_{\theta}$
\ENDWHILE
\STATE Deduplicate $\mathcal{F}_{\mathrm{paths}}$ by $(M,B)$ into
$\mathcal{F}(Q)$, keeping the best path score $\bar{L}_{\theta}(Q,A)$ of each configuration
\IF{a critic is available}
    \STATE $S_{\mathrm{final}}(Q,A)\leftarrow\widehat{U}_{\psi}(Q,A)$ for each $A\in\mathcal{F}(Q)$
\ELSE
    \STATE $S_{\mathrm{final}}(Q,A)\leftarrow
    \bar{L}_{\theta}(Q,A)-\lambda_{\mathrm{len}}|B|$
\ENDIF
\RETURN $\operatorname{Top}_{K}(\mathcal{F}(Q);S_{\mathrm{final}}(Q,\cdot))$
\end{algorithmic}
\end{algorithm}

\section{Experimental Setup Details}
\label{app:setup_details}

\subsection{Datasets}
\label{app:dataset_details}

\noindent\textbf{AgentSelect.} AgentSelect aggregates over 40 sources: 111,179 queries, 107,721 agents, and 251,103 interactions, with a catalog of 231 LLM-only entries and 12,099 tools. We evaluate on all 629 AgentSelect test queries.

\noindent\textbf{SkillsBench.} We sample 1,000 successful SkillsBench trajectories with verifier reward 1.0, which give 97 queries across 79 tasks and 615 query--agent pairs. The inventory contains 19 LLM identifiers, 155 skills, and 20 tool or action components. For SkillsBench, we split by task, training on 16 tasks (19 queries) and evaluating on the remaining 63 tasks (78 queries). Appendix~\ref{app:skillsbench_data} provides further details on dataset construction.

\noindent\textbf{Execution datasets.} Three further datasets record the execution outcome of each configuration. \textbf{AgentSelect-Exec} attaches execution feedback to 506 AgentSelect queries: for each query, we execute the ten configurations our method ranks highest, calling the selected backbone, simulating tool responses with an LLM, and scoring each run with a blinded LLM judge on completion (0--4), correctness (0--4), and constraint adherence (0--2). Building it consumes about 50.4M tokens, and Appendix~\ref{app:agentselect_exec} details the construction and cost. \textbf{BFCL}~\cite{BFCL} and \textbf{SWE-Skills-Bench}~\cite{SWESkillsBench} release execution results for fixed configurations, which we use as native rewards, with each task forming one query. From BFCL we take the 240 executable Python tasks, each run by 18 backbones with the task's function menu. The reward is 1 if the calls pass the execution check and 0 otherwise. SWE-Skills-Bench (v2) contains 980 software-engineering tasks over 98 skills, each run under eight configurations, namely four model--harness pairs with the task's skill enabled or disabled. The reward is the fraction of hidden tests passed. We split both benchmarks by task into training, validation, and test tasks, 161/35/44 on BFCL and 690/60/230 on SWE-Skills-Bench, where the ten tasks of a skill stay in one split. Appendix~\ref{app:execution_benchmarks} details both benchmarks, Section~\ref{sec:execution_adaptation} uses AgentSelect-Exec, and Section~\ref{sec:cross_domain_transfer} uses BFCL and SWE-Skills-Bench.

\subsection{Implementation}
\label{app:implementation_details}

\noindent \textbf{Retrieval.}
On AgentSelect, we concatenate CF LLM, CF Historical Bundle, and Semantic Individual Tool Retrieval outputs as separate context sections, preserving bundle structure and shortening component profiles to fit the input budget. Text2Bundle and subsequent generative methods, including ours, share this evidence. Component-Ret uses LLM and individual-tool outputs, and Bundle-Ret uses LLM and historical-bundle outputs.

\noindent \textbf{Llama Fine-Tuning.}
We fine-tune Llama-3-8B on AgentSelect for one epoch using LoRA rank 16, scaling factor 32, dropout 0.05, learning rate $10^{-4}$, and effective batch size 32. We update LoRA parameters, token embeddings, and the output head, with a maximum sequence length of 8,192 tokens. Loss covers only configuration output tokens. For SkillsBench, we further fine-tune the AgentSelect model on the 109 training configurations for three epochs with learning rate $5\times10^{-5}$ and effective batch size 8, updating only LoRA parameters while freezing embeddings and the output head. Generator scores rank the SkillsBench configurations in Table~\ref{tab:skillsbench_results} (Section~\ref{subsec:bundle_generation}). Section~\ref{sec:cross_domain_transfer} uses a few-shot protocol.

\noindent \textbf{Critic Training.}
We initialize the critic from AgentEmbedding~\cite{AgentEmbedding}, using 1,000 AgentSelect training queries and 150 held-out validation queries. Stage 1 uses four epochs of teacher-supervised pairwise ranking plus a pointwise loss weighted by 0.1, and we select the checkpoint with the highest validation NDCG@10. Stage 2 refines with preference pairs for redundant additions, missing tools, and replacements, retaining teacher comparisons between LLMs. It uses a 0.1 pairwise margin without pointwise loss, and we select its checkpoint on the validation queries. Both stages update the encoder and scoring head with AdamW, using learning rates $(5\times10^{-6},10^{-4})$ and $(2\times10^{-6},2\times10^{-5})$ and effective batch sizes of 32 and 24 pairs, respectively. Stage~3 continues from the Stage~1+2 critic on the AgentSelect-Exec training split. It minimizes $\mathcal{L}_{\mathrm{critic}}^{(3)}$ (Section~\ref{subsec:bundle_preference}). Training runs for four epochs with AdamW (weight decay 0.01), learning rates $(2\times10^{-6},10^{-4})$, and four queries per batch, selecting the checkpoint with the lowest validation loss. For BFCL and SWE-Skills-Bench, target fine-tuning likewise starts from the Stage~1+2 critic, trains for three epochs on each fraction of the training tasks with a task-balanced loss on native rewards, and selects the checkpoint on the validation tasks.

\noindent \textbf{Search Configuration.}
On AgentSelect, we use validation-selected adaptive beam bounds $(b_{\min},b_{\max})=(1,2)$, together with retain ratio $r_{\mathrm{keep}}=0.4$ in Eq.~\eqref{eq:dynamic_width}. We further set the proposal budget $P=5$, $L_{\max}=6$, and $K=10$. Retrieval supplies at most 10 backbones in $\mathcal{M}_r(Q)$. On both AgentSelect and SkillsBench, the no-tool penalty is $\lambda_{\emptyset}=10$. For example, when retrieval supplies 10 backbones, the generator proposes 5 roots and pruning keeps $W(5)=2$ of them. In general, these settings give $W(\eta)=1$ for $\eta\leq2$ and $W(\eta)=2$ otherwise, so at most two paths remain active after each pruning step.

\section{Construction of the SkillsBench Recommendation View}
\label{app:skillsbench_data}

\noindent\textbf{Trace extraction.}
We use runs from the public SkillsBench~\cite{SkillsBench} leaderboard that
use skills and pass the verifier with reward 1.0, and sample 1,000 of
them (Appendix~\ref{app:dataset_details}).
Each trace gives its backbone in the result record. A skill counts as used
when the trace activates it or reads or executes its files. Tools keep their
observed names, and harness actions form a separate category. The used skills
and tools form the component set.

\noindent\textbf{Queries and reference configurations.}
We group traces by source task and normalized initial query, merge duplicate
model--component-set profiles, and retain up to ten distinct successful
configurations per query. All references of a query have the
same verifier reward and are unordered.
Evaluation matches each candidate against any individual reference
(Appendix~\ref{app:evaluation_metrics}). All queries of a task fall in one
split. The main recommendation
experiment uses the train/test split in Table~\ref{tab:dataset_statistics};
Section~\ref{sec:cross_domain_transfer} uses a separate few-shot protocol.

\noindent\textbf{Inventory.}
The inventory contains the observed components and backbones. Each skill is
described by its public skill document. Methods receive the query text and
the descriptions of all catalog components.

\section[Construction of AgentSelect-Exec]{Construction of AgentSelect-Exec}
\label{app:agentselect_exec}

AgentSelect-Exec attaches execution feedback to AgentSelect queries. We build it in four steps and report its token cost.

\noindent\textbf{Candidate configurations.}
We take 506 AgentSelect queries. The backbones are 18 LLMs:
GPT-6-Astra, GPT-6-Luna, GPT-6.1-Sol, GPT-6.1-Terra, DeepSeek-R1,
DeepSeek-V3.2, DeepSeek-V4-Pro, DeepSeek-V4.1-Flash, Qwen3-8B, Qwen3-32B,
Qwen3.6-35B-A3B, Qwen3.8-27B, Kimi-K2.6, GLM-4.5-Air, GLM-5.3,
Seed-OSS-36B-Instruct, LongCat-2.0, and Hy4-preview. For each query, our method
(Section~\ref{sec:method}) ranks configurations over these backbones with the
Stage~1+2 critic (Section~\ref{subsec:bundle_generation}), and we keep the ten highest-ranked
configurations, 5,060 in total.

\noindent\textbf{Execution.}
Each configuration is executed once. The harness calls the selected backbone through
its provider API for at most 12 assistant turns and 24 tool calls,
with up to 4,000 output tokens per turn. The configuration's tools are exposed
with their documented schemas. A GPT-5.6-Luna simulator with a
per-configuration state answers every tool call.

\noindent\textbf{Judging.}
A GPT-5.6-Terra judge with low reasoning effort scores each run on completion
(0--4), correctness (0--4), and constraint adherence (0--2). The judge
sees the user request, the final answer, the observed evidence, submitted artifacts,
and the termination status, and is blind to the recommending method, the backbone
identity, and the rank. The mean total score over all runs is 7.38 out of 10.

\noindent\textbf{Splits.}
We split the queries at random into 200 training, 106 validation,
and 200 test queries, each with ten executed configurations
(Table~\ref{tab:dataset_statistics}).

\noindent\textbf{Cost.}
Table~\ref{tab:agentselect_exec_tokens} reports the token usage, about 50.4M
in total.

\begin{table}[h]
\centering
\caption{Token usage for building AgentSelect-Exec.}
\label{tab:agentselect_exec_tokens}
\small
\begin{tabular}{@{}llrrr@{}}
\toprule
Role & Model & Input tokens & Output tokens & Total tokens \\
\midrule
Backbone & 18 LLMs & 19,777,574 & 4,413,132 & 24,190,706 \\
Tool simulator & GPT-5.6-Luna & 12,870,440 & 1,622,589 & 14,493,029 \\
Judge & GPT-5.6-Terra & 10,206,606 & 1,496,896 & 11,703,502 \\
\midrule
Total & & 42,854,620 & 7,532,617 & 50,387,237 \\
\bottomrule
\end{tabular}
\end{table}

\section[BFCL and SWE-Skills-Bench]{BFCL and SWE-Skills-Bench}

\label{app:execution_benchmarks}

Both benchmarks release execution results for a fixed set of configurations on every task. We treat each task as one query, its executed configurations as the candidate pool, and the released outcome of each run as its native reward. We split each benchmark by task into training, validation, and test tasks (Table~\ref{tab:dataset_statistics}).

\noindent\textbf{BFCL.}
BFCL~\cite{BFCL} evaluates function calling: given a user request and function
documentation, a model selects the functions to call and fills in their arguments,
and each call is checked either against the expected structure or by executing it.
Its first release contains 2,000 question--function--answer examples across 40
application domains. We use its 240 executable Python tasks: 100 single-function,
50 multiple-function (choosing among several functions), 50 parallel-call, and
40 parallel multiple-function tasks. We take the released results of 18 models:
Claude-3-Haiku, Claude-3-Sonnet, Claude-3-Opus,
Claude-3.5-Sonnet, Command-R+, FireFunction-v1 and v2, GPT-3.5-Turbo-0125,
GPT-4-0613, GPT-4-1106-Preview, GPT-4-0125-Preview, GPT-4-Turbo-2024-04-09,
GPT-4o-2024-05-13, GPT-4o-2024-08-06, GPT-4o-mini, Functionary-Small-v3.1 and v3.2,
and Functionary-Medium-v3.1, giving 4,320 runs. Every model receives the task's
full function menu, so a task's configurations differ only in the backbone.
A run's reward is 1 if its calls pass the execution check and 0 otherwise.
The 161/35/44 train/validation/test tasks hold 2,898/630/792 runs.

\noindent\textbf{SWE-Skills-Bench.}
SWE-Skills-Bench~\cite{SWESkillsBench} measures how much an agent skill helps in
software engineering. Each skill is paired with a real code repository and ten
requirement tasks, whose acceptance criteria are converted into hidden executable
tests. Each task is run twice under the same repository, environment, and model
configuration, once without the skill and once with the designated skill loaded.
We use version~2 and its released results, which cover 98 skills and 980 tasks
under four model--harness pairs: Codex with GPT-5.4-mini, Codex with Kimi-K2.6,
Claude Code with Kimi-K2.6, and Claude Code with Claude Haiku~4.5. The skills span
API development (26), developer tools (18), data science and machine learning
(17), deployment and operations (14), analytics and monitoring (12), and
security and testing (11). Each task thus has eight configurations, and each of the
7,840 runs is rewarded by the fraction of hidden tests passed, in $[0,1]$. All ten
tasks of a skill fall into the same split, giving 69/6/23 skills,
690/60/230 tasks, and 5,520/480/1,840 runs for training, validation,
and test. On both benchmarks, RAG-Qwen3.8-27B receives the task and the full descriptions of all candidate resources and returns a strict ranking of the executed pool.

\section{Baseline Adaptations and Output Protocols}
\label{app:baselines}

\subsection{Retrieval-Based Composition}
\label{app:retrieval_baselines}

\noindent\textbf{Component-Ret.}
We independently retrieve LLMs via CF and tools via semantic retrieval.
The top-10 LLMs are paired with the same six
highest-ranked tools, forming ten configurations ranked by LLM CF score.

\noindent\textbf{Bundle-Ret.}
Separate CF models retrieve one top-ranked LLM and ten historical tool
bundles. Each intact bundle is paired with that LLM and ranked by bundle
CF score.

\subsection{Text2Bundle}
\label{app:text2bundle}

We adapt Text2Bundle~\cite{Text2Bundle} by mapping items to tools and using
the query as the request. We omit user-ID embeddings and personalization
rewards. Queries and tool
descriptions use AgentEmbedding~\cite{AgentEmbedding}, an EasyRec~\cite{easyrec}
model tuned for agent-resource retrieval. We retain the Transformer state
encoder, action-relation modeling, and learned termination action.

Pairwise bundle-completion pretraining precedes Double-DQN training with
experience replay. The action pool deduplicates tools from retrieved
historical bundles and semantic retrieval. Missing target tools are added
to the pool during training, and validation and inference use the retrieved
candidates.

The CF-selected LLM stays fixed while beam search extends the single-output
policy to Top-$K$ tool sets. Paths are ranked by negative cumulative action
regret: each increment is the chosen action's Q-value minus the highest
valid-action Q-value. The greedy path scores zero and comes first.
Duplicate tool sets keep their best path, and a size cap ends search.

\subsection{DDBC-Adapt}
\label{app:ddbc}

We adapt DDBC~\cite{tu2026discrete}, which uses masked discrete diffusion
and residual quantization. DDBC-Adapt operates over a component vocabulary
built from the training split.
Frozen MPNet embeddings of LLM and tool descriptions, PCA, and residual
$k$-means replace its representation learner and quantizer. An extra code
separates components with identical quantized codes.

A query-conditioned denoiser jointly generates one LLM and a tool set with
typed boundaries. Training masks component codes, preserves structural
tokens, and randomizes tool order. Decoding retains absorbing masks, enforces code
compatibility, and excludes duplicate tools, so it can generate new
combinations of the components in its training vocabulary.

Task-trained semantic adapters estimate query--component relevance over the
training vocabulary. Relevance
probabilities are summed over legal components sharing each code value, and
the resulting log masses augment denoiser logits. Distinct valid
configurations are ranked by LLM and summed tool relevance, weighted by the
predicted probability of their tool count.

\subsection[Adapted Tool Retrieval and LLM Routing]{Adapted Tool Retrieval and LLM Routing}
\label{app:adapted_baselines}

ToolGen~\cite{ToolGen}, ToolWeaver~\cite{ToolWeaver}, PORTS~\cite{PORTS}, AnyTool~\cite{du2024anytool}, and RouteLLM~\cite{RouteLLM} each address one part of agent resource use and do not output a complete configuration. ToolGen selects the next tool for a fixed backbone within its agent loop, ToolWeaver generates tools as hierarchical code sequences that capture tool co-usage, PORTS aligns a dense tool retriever with LLM preferences over tool selection, AnyTool collects individual APIs through hierarchical LLM agents, and RouteLLM routes a query between a strong and a weak LLM without tools. We adapt each method to produce configurations while keeping its core mechanism, and pair its output with the remaining module of Component-Ret.

\noindent\textbf{ToolGen.} Following ToolGen, we first train tool-token memorization on 47,800 document-to-token examples, one per tool in our catalog. We then fine-tune the model to generate the complete tool sequence of each training configuration. Beam search produces candidate tool sets, each paired with the CF backbone ranking, and the fusion weight of tool and backbone scores is selected on 616 validation queries.

\noindent\textbf{ToolWeaver.} We encode each tool document with a frozen MPNet encoder and quantize it with a residual-quantized autoencoder into two code levels of 1,024 entries each. A graph term in the quantizer pulls together tools that co-occur in training configurations, so co-usage shapes the codes, and every catalog tool keeps a unique two-token code. The language model starts from the ToolGen memorization checkpoint, and each new code token is initialized from the memorized tokens of the tools it covers. With LoRA, we first align queries with the codes of single tools from the training configurations and then fine-tune the model to generate the codes of complete training bundles. Beam search over valid codes produces bundles of one to six tools, which are combined with the CF backbone ranking as in ToolGen, with weights selected on validation queries.

\noindent\textbf{PORTS.} We split each training configuration into query--tool positives, and a frozen Meta-Llama-3-8B scores the identifier of the positive tool when conditioned on each candidate tool document. A BGE-base encoder, shared by queries and tool documents, learns to match this preference distribution over the positive and three hard negatives, together with an odds-ratio preference term that ranks the positive above each negative. Hard negatives are the encoder's nearest non-positive tools and are refreshed during training. At test time, the top two retrieved tools form the bundle, and its size and the checkpoint are selected by tool F1 on validation queries. The bundle is paired with the CF backbone ranking.

\noindent\textbf{AnyTool.} We keep AnyTool's category--service--API navigation and solvability check, use the collected APIs as the tool set, and pair it with the CF backbone ranking.

\noindent\textbf{RouteLLM.} We replace the backbone ranking of Component-Ret with RouteLLM's matrix-factorization (MF) router, trained on backbone positives with TF-IDF query features, and keep the six semantically retrieved tools.

Table~\ref{tab:adapted_baselines} lists each method's tool and backbone sources, and Table~\ref{tab:main_results} gives the results.

\begin{table}[h]
\centering
\caption{Sources of tools and backbones in the adapted baselines.}
\label{tab:adapted_baselines}
\small
\begin{tabular}{@{}lll@{}}
\toprule
Method & Tools from & Backbones from \\
\midrule
Component-Ret & semantic retrieval & CF ranking \\
PORTS & LLM-aligned dense retrieval & CF ranking \\
AnyTool & hierarchical API retrieval & CF ranking \\
RouteLLM & semantic retrieval & MF router \\
ToolGen & generated tool sequence & CF ranking \\
ToolWeaver & generated hierarchical tool codes & CF ranking \\
\midrule
Ours & joint generation & joint generation \\
\bottomrule
\end{tabular}
\end{table}

\section{Evaluation Metric Definitions}
\label{app:evaluation_metrics}

Each method returns up to ten ranked configurations per query. Let
$(M_r,B_r)$ be the backbone and tool set at rank $r$, and $(M^\ast,B^\ast)$ a
reference configuration. The backbone and each tool count as one component
(on SkillsBench, tools include skills and actions). A query can have several
references, and every metric uses the reference that a candidate matches best.
The component recall $c_r$ and precision $p_r$ at rank $r$ give the
recall-based metrics and the rank-discounted scores:
\begin{gather*}
c_r=\frac{\mathbb{I}[M_r=M^\ast]+|B_r\cap B^\ast|}{1+|B^\ast|},\qquad
p_r=\frac{\mathbb{I}[M_r=M^\ast]+|B_r\cap B^\ast|}{1+|B_r|},\\[2pt]
\mathrm{ToolR@1}=\frac{|B_1\cap B^\ast|}{|B^\ast|},\qquad
\mathrm{CompR@1}=c_1,\\[2pt]
\mathrm{RDCR@10}=\frac{\sum_{r=1}^{10}w_r\,c_r}{\sum_{r=1}^{10}w_r},\qquad
\mathrm{RDCP@10}=\frac{\sum_{r=1}^{10}w_r\,p_r}{\sum_{r=1}^{10}w_r},\qquad
w_r=\frac{1}{\log_2(r+1)}.
\end{gather*}
Missing ranks count as zero. Extra tools do not lower RDCR but lower RDCP.

The complete-recovery metrics ask whether a reference is fully recovered.
Tool-Hit@10 is 1 if any of the top ten configurations contains all tools of
a reference, regardless of the backbone. CR-Hit@$k$ also requires the reference
backbone and checks the top $k$ configurations ($k=1,10$). CR-MRR@10 is the
reciprocal rank of the first such hit, and 0 if there is none in the top ten.
All of them allow extra tools. Every score is averaged over queries and
reported on a 0--100 scale.

\section{Retrieval Analysis}
\label{app:retrieval_analysis}

These studies measure the coverage of each retrieval branch in
Appendix~\ref{app:retrieval_design}.

\noindent\textbf{Evaluation protocol.}
Tool retrieval uses 1,000 unique AgentSelect Part~II queries sampled
at random. Macro recall averages the fraction of reference tools
retrieved per query, and the all-covered rate requires their complete coverage.
For historical bundles, these measures use the merged tool set. Exact hit
requires one retrieved bundle to equal the reference tool set, while complete
hit permits additional tools. The cutoff $K$ counts tools or bundles,
respectively, and Avg.\ Tools counts the unique retrieved tools after merging.
LLM retrieval uses all 23,073 unique Part~I queries, each with multiple
ranked reference LLMs. Precision (P), recall (R), MRR, and MAP use binary
relevance. Hit requires any relevant LLM, and the all-covered rate (All)
requires every reference LLM. For nDCG, reference rank $r$ receives gain
$1/\log_2(r+1)$.

\begin{wraptable}{r}{0.44\linewidth}
\vspace{-4mm}
\centering
\small
\caption{Individual-tool retrieval (\%).}
\label{tab:tool_level_retrieval}
\setlength{\tabcolsep}{6pt}
\begin{tabular}{@{}ccc@{}}
\toprule
$K$ & Macro Recall & All Covered \\
\midrule
5  & 39.95 & 21.10 \\
10 & 48.30 & 30.20 \\
15 & 51.93 & 34.00 \\
20 & 55.74 & 38.00 \\
50 & 65.46 & 50.60 \\
\bottomrule
\end{tabular}
\vspace{-3mm}
\end{wraptable}
\noindent\textbf{Tool retrieval.}
Individual-tool retrieval covers few reference tools: increasing $K$ from 5
to 50 raises macro recall from 39.95\% to 65.46\% and the all-covered
rate from 21.10\% to 50.60\% (Table~\ref{tab:tool_level_retrieval}).
Historical bundles cover far more. At $K=5$, they yield 5.52 unique tools on
average and 88.09\% macro recall (Table~\ref{tab:tool_bundle_retrieval}).
Merging adds only 0.50 percentage points of complete coverage beyond a single
bundle's complete hit. Increasing $K$ to 20 raises the merged all-covered rate
to 90.00\%, with 17.21 tools on average.

\noindent\textbf{LLM retrieval.}
At $K=10$, CF LLM retrieval has a 97.95\% hit rate and 0.8565 MRR
(Table~\ref{tab:cf_llm_retrieval}). Increasing $K$ to 50 mainly recovers
additional relevant models: recall rises from 57.07\% to 92.34\%,
while the hit rate rises to 99.91\%.

\begin{table}[H]
\centering
\small
\caption{CF tool-bundle retrieval (\%, except Avg.\ Tools).}
\label{tab:tool_bundle_retrieval}
\begin{tabularx}{\linewidth}{@{}c*{5}{>{\centering\arraybackslash}X}@{}}
\toprule
$K$ & Exact Hit & Complete Hit & Macro Recall & All Covered & Avg.\ Tools \\
\midrule
5  & 81.20 & 85.10 & 88.09 & 85.60 & 5.52 \\
10 & 85.00 & 87.60 & 89.49 & 88.20 & 9.34 \\
15 & 86.70 & 88.50 & 90.10 & 89.20 & 13.19 \\
20 & 88.20 & 89.30 & 90.69 & 90.00 & 17.21 \\
\bottomrule
\end{tabularx}

\vspace{4mm}
\caption{CF LLM retrieval (P, R, Hit, and All in \%).}
\label{tab:cf_llm_retrieval}
\begin{tabularx}{\linewidth}{@{}c*{7}{>{\centering\arraybackslash}X}@{}}
\toprule
$K$ & P & R & Hit & All & MRR & MAP & nDCG \\
\midrule
5  & 69.43 & 35.43 & 94.75 & 0.00  & 0.8521 & 0.6387 & 0.6327 \\
10 & 56.21 & 57.07 & 97.95 & 3.61  & 0.8565 & 0.4877 & 0.6244 \\
15 & 45.30 & 68.81 & 98.99 & 13.15 & 0.8574 & 0.5519 & 0.6744 \\
20 & 37.57 & 76.01 & 99.45 & 23.00 & 0.8576 & 0.5816 & 0.7011 \\
50 & 18.29 & 92.34 & 99.91 & 64.63 & 0.8578 & 0.6250 & 0.7512 \\
\bottomrule
\end{tabularx}
\end{table}

\section{Sensitivity to Beam Bounds and Critic Weight}
\label{app:beam_critic_sensitivity}

\noindent\textbf{Validation protocol.}
With the generator and critic of Table~\ref{tab:main_results}, we test adaptive beam
bounds $(b_{\min},b_{\max})$ of $W(\eta)$ in Eq.~\eqref{eq:dynamic_width} over $\{(1,2),(2,4),(3,5),(4,7),(6,10)\}$ and critic weights
$\alpha_c\in\{0,0.5,1.0,1.5,2.0\}$ on validation queries, holding other settings fixed.
To see how much the critic should count, this study ranks the final
candidates by a weighted sum of the critic score (weight $\alpha_c$) and the
generator score (weight 0.15), both standardized within the candidate pool,
minus 0.1 per tool. At $\alpha_c=0$, ranking uses generator scores alone,
and a large $\alpha_c$ approaches ranking by the critic alone.

\begin{figure}[H]
\centering
\includegraphics[width=0.62\textwidth]{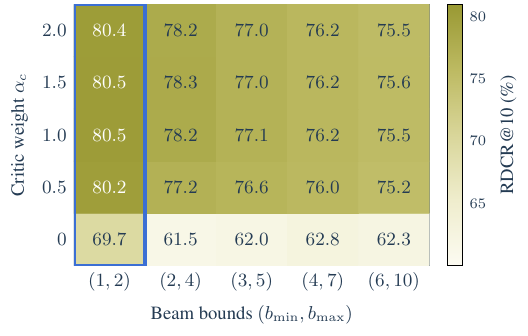}
\caption{Validation RDCR@10 over beam bounds and critic weights. The outlined column is the bounds (1, 2) that we use.}
\label{fig:beam_critic_stagewise}
\end{figure}

\noindent\textbf{Beam bounds.} Beam bounds matter more than the critic weight. Bounds (1, 2) give the highest RDCR@10 at every tested weight (Figure~\ref{fig:beam_critic_stagewise}). Without the critic, widening the bounds from (1, 2) to (4, 7) lowers RDCR@10 from 69.7\% to 62.8\%. With the critic, RDCR@10
falls by about 5 points from (1, 2) to (6, 10), so we use (1, 2).

\noindent\textbf{Critic weight.} Critic reranking raises RDCR@10 at every beam setting, e.g., from 69.7\% to 80.2\% at (1, 2) and from 62.8\% to 76.0\% at (4, 7). Once the critic is used, its weight matters little. Across positive
weights, RDCR@10 varies by at most 1.1 points per beam setting, and at
(1, 2) it spans 80.2--80.5\%. The largest weight,
$\alpha_c=2.0$, where the critic dominates the ranking, reaches 80.4\%,
on par with $\alpha_c=0.5$. Generator scores therefore add little once the
critic ranks the candidates, and our final ranking uses the critic alone
(Section~\ref{subsec:bundle_generation}).

\section[Output Validity of Constrained Decoding]{Output Validity of Constrained Decoding}
\label{app:decoding_validity}

Table~\ref{tab:decoding_validity} compares single outputs of free and constrained single-path decoding.

\begin{wraptable}{r}{0.5\linewidth}
\vspace{-4mm}
\centering
\captionsetup{font=footnotesize,skip=3pt}
\caption{Decoding validity (\%).}
\label{tab:decoding_validity}
\footnotesize
\setlength{\tabcolsep}{3.2pt}
\begin{tabular}{@{}lcccc@{}}
\toprule
& \multicolumn{2}{c}{AgentSelect} & \multicolumn{2}{c}{SkillsBench} \\
\cmidrule(lr){2-3}\cmidrule(l){4-5}
& Free & Constr. & Free & Constr. \\
\midrule
Valid & 97.0 & 100.0 & 48.7 & 100.0 \\
\midrule
Incomplete & 0.0 & 0.0 & 3.8 & 0.0 \\
Malformed & 0.0 & 0.0 & 43.6 & 0.0 \\
Unknown name & 3.0 & 0.0 & 3.8 & 0.0 \\
\midrule
Same config. & \multicolumn{2}{c}{92.0} & \multicolumn{2}{c}{39.7} \\
\bottomrule
\end{tabular}
\vspace{-3mm}
\end{wraptable}
\noindent An output is valid if it ends within 256 new tokens, follows the configuration format, and names only catalog components. On SkillsBench, the AgentSelect-trained generator is applied without target training. Same configuration is the share of queries on which both decodings return the same backbone and tool set. Free generation is not always valid. On AgentSelect, 3\% of its outputs name components outside the catalog. On SkillsBench, about half of its outputs are invalid, mostly malformed, because the generator has not seen the target format. Constrained decoding keeps every output valid in both settings, and the two decodings agree far less often on SkillsBench than on AgentSelect.

\section{Online Recommendation and Execution}
\label{app:e2e_execution}
\label{sec:e2e}
\label{app:execution_baselines}

This section details the study of
Section~\ref{sec:online_execution}. Appendices~\ref{app:e2e_setup}
and~\ref{app:e2e_runtime} give the setup. Appendices~\ref{app:execution_baselines_skillsbench}
and~\ref{app:e2e_native} cover SkillsBench: the online execution of the
top-ranked configurations, and an offline ranking of the official SkillsBench
runs. Appendix~\ref{app:execution_baselines_agentselect} covers AgentSelect,
including our full top-10 lists.

\subsection{Queries, resources, and candidate generation}
\label{app:e2e_setup}
\label{app:e2e_generation}

The study uses 25 AgentSelect test queries and five SkillsBench test queries.
The SkillsBench queries cover travel planning, dialogue parsing, invoice fraud
detection, bike rebalancing, and software dependency auditing.

All queries within a dataset share one resource inventory
(Table~\ref{tab:e2e_inventory}).

\begin{table}[H]
\centering
\caption{Resources and runtime budgets of the online study.}
\label{tab:e2e_inventory}
\small
\setlength{\tabcolsep}{4pt}
\begin{tabularx}{\linewidth}{@{}Xrr@{}}
\toprule
Setting & AgentSelect & SkillsBench \\
\midrule
Queries & 25 & 5 \\
Backbone candidates & 14 & 14 \\
API endpoints / skills & 66 / 16 & 66 / 16 \\
Maximum components & 6 & 10 \\
Assistant turns / tool calls & 12 / 24 & 24 / 48 \\
Output tokens per response & 4,000 & 6,000 \\
\bottomrule
\end{tabularx}
\end{table}

All 14 backbone models form the candidate set $\mathcal{M}_r(Q)$, and the
generator keeps four root proposals. Each root receives one component
expansion before the first pruning step, and each active path considers up to
ten legal component proposals. Pruning uses the settings of
Section~\ref{sec:implementation}.

On AgentSelect, the Stage~3 critic ranks the completed configurations.
On SkillsBench, the generator score $\bar L_{\theta}$ ranks them. Each query receives ten
configurations.

\subsection{Execution and scoring}
\label{app:e2e_runtime}
\label{app:e2e_evaluation}

Of the 66 APIs, 11 call public services or a local adapter, and the other 55
are simulated by GPT-5.6-Luna following MirrorAPI~\cite{MirrorAPI}. The
simulator sees the endpoint description, schema, and arguments, and is blind
to the recommendation rank, score, and backbone. Each configuration runs at temperature
zero within the budgets of Table~\ref{tab:e2e_inventory}.

On SkillsBench, the selected skills are mounted in network-isolated containers,
and the original hidden tests check the exported artifacts. On AgentSelect, the
selected skills are given to the backbone as textual instructions.

GPT-5.6-Terra judges each run from the user request, the outputs, the tool
observations, and any verifier results, blind to the rank, scores, and
backbone. It assigns 0--4 points for task completion, 0--4 for
correctness, and 0--2 for constraint satisfaction.

\subsection{SkillsBench with Native Verifiers}
\label{app:execution_baselines_skillsbench}

\noindent\textbf{Setting.} The five tasks are those of Appendix~\ref{app:e2e_setup}. The baseline is RAG-Qwen3.8-27B. Each method recommends an execution LLM and a set of skills. Ours selects Qwen3.8-27B for all five tasks, and RAG-Qwen3.8-27B selects DeepSeek-V4-Pro. Two tests follow. The first executes each recommended configuration once. The second runs the skills of each method three times with a fixed LLM backbone, GPT-5.6-Luna, which isolates skill selection. The native verifier decides pass or fail, and GPT-5.6-Terra scores each run with the rubric of Appendix~\ref{app:e2e_evaluation}. Execution tokens are counted in the first test.

\begin{table}[h]
\centering
\caption{Execution on SkillsBench.}
\label{tab:execution_baselines_skillsbench}
\small
\setlength{\tabcolsep}{3.3pt}
\arrayrulecolor{TableOneInk!65}
\begin{tabular}{@{}l*{10}{c}@{}}
\toprule
& \multicolumn{6}{c}{\makecell{Recommended configuration\\(one run)}} & \multicolumn{4}{c}{\makecell{Fixed LLM backbone\\(three runs)}} \\
\cmidrule(lr){2-7}\cmidrule(l){8-11}
& \multicolumn{2}{c}{Verifier} & \multicolumn{2}{c}{Judge (0--10)} & \multicolumn{2}{c}{Tokens (K)} & \multicolumn{2}{c}{Passes} & \multicolumn{2}{c}{Judge (0--10)} \\
\cmidrule(lr){2-3}\cmidrule(lr){4-5}\cmidrule(lr){6-7}\cmidrule(lr){8-9}\cmidrule(l){10-11}
Task & Ours & RAG & Ours & RAG & Ours & RAG & Ours & RAG & Ours & RAG \\
\midrule
Travel planning   & \ding{51} & \ding{51} & \textbf{8.0}  & 7.5  & \textbf{205} & 417 & 0/3 & 0/3 & \textbf{5.80}  & 5.17 \\
Dialogue parsing  & \ding{51} & \ding{51} & 10.0 & 10.0 & \textbf{76}  & 342 & \textbf{3/3} & 2/3 & \textbf{10.00} & 9.03 \\
Invoice fraud     & \ding{51} & \ding{55} & \textbf{10.0} & 8.0  & \textbf{118} & 135 & \textbf{3/3} & 0/3 & \textbf{10.00} & 8.67 \\
Bike rebalancing  & \ding{55} & \ding{55} & \textbf{5.0}  & 0.0  & 688 & \textbf{569} & 0/3 & 0/3 & \textbf{4.50}  & 3.17 \\
Dependency audit  & \ding{51} & \ding{51} & 10.0 & 10.0 & \textbf{47}  & 129 & 3/3 & 3/3 & \textbf{9.83}  & 9.33 \\
\midrule
All tasks & \textbf{4/5} & 3/5 & \textbf{8.6} & 7.1 & \textbf{1,134} & 1,593 & \textbf{9/15} & 5/15 & \textbf{8.03} & 7.07 \\
\bottomrule
\end{tabular}
\end{table}

\noindent\textbf{Results.} Table~\ref{tab:execution_baselines_skillsbench} gives the per-task results, where RAG denotes RAG-Qwen3.8-27B and bold marks the better value of each pair. With the recommended configurations, ours passes the verifier on four of five tasks and RAG-Qwen3.8-27B on three. Ours also has a higher mean judge score (8.6 versus 7.1) and uses 29\% fewer tokens. On invoice fraud detection, only ours passes. On bike rebalancing, both fail, and ours completes part of the task (5.0 versus 0.0). With the fixed LLM backbone, our skills pass 9 of 15 runs, compared with 5, and score higher on all five tasks (8.03 versus 7.07).

\subsection[Native execution on official SkillsBench runs]{Native execution on official SkillsBench runs}
\label{app:e2e_native}

\noindent\textbf{Data.} We use the official SkillsBench runs released with the benchmark, executed in OpenHands with three backbones (Opus~4.8, Gemini~3.5 Flash, and MiniMax~M3), each with and without the task's skills. We split the tasks into 36 training, 14 validation, and 21 test tasks, and report results on the test tasks. A configuration is a backbone with or without skills; its reward is the mean hidden-test reward over its 1--3 repeated runs (93 configurations, 255 runs). The Stage~1+2 critic, trained on AgentSelect, scores each configuration from the task request, the backbone, and the skills.

\noindent\textbf{Baselines and metrics.} The backbone prior ranks configurations by their backbone's mean reward on the training tasks; the skill prior ranks skill-enabled configurations first. Top-1 is the reward of the top-ranked configuration, averaged over tied configurations. Pairwise concordance counts correctly ordered within-task pairs with different rewards, with predictor ties counted as 0.5; same-backbone pairs differ only in whether skills are enabled, and in both-succeed pairs both configurations receive a nonzero reward. nDCG@3 uses linear gains. Intervals are 95\% bootstrap intervals over tasks.

\begin{table}[H]
\centering
\caption{Official SkillsBench test runs (rewards and concordance in \%).}
\label{tab:e2e_native}
\small
\setlength{\tabcolsep}{1.6pt}
\begin{tabular}{@{}lccccccc@{}}
\toprule
& & & \multicolumn{3}{c}{Pairwise concordance} & \\
\cmidrule(lr){4-6}
Ranker & Top-1 reward & nDCG@3 & All & Same backbone & Both succeed & Spearman $\rho$ \\
\midrule
Random & 41.9 {\scriptsize[28.4, 56.2]} & -- & 50.0 & 50.0 & 50.0 & 0.00 \\
Backbone prior & 45.2 {\scriptsize[26.9, 63.5]} & 67.3 & 56.4 & 50.0 & 66.2 & 0.16 \\
Skill prior & 49.0 {\scriptsize[35.2, 63.2]} & 81.9 & 61.6 & 68.0 & 56.8 & 0.29 \\
Ours (Stage~1+2 critic) & \textbf{54.0} {\scriptsize[36.5, 69.8]} & \textbf{82.1} & \textbf{69.8} & \textbf{72.0} & \textbf{73.0} & \textbf{0.36} \\
\midrule
Oracle & 72.1 {\scriptsize[57.8, 84.8]} & 100.0 & 100.0 & 100.0 & 100.0 & -- \\
\bottomrule
\end{tabular}
\end{table}

\begin{figure}[H]
\centering
\includegraphics[width=\linewidth]{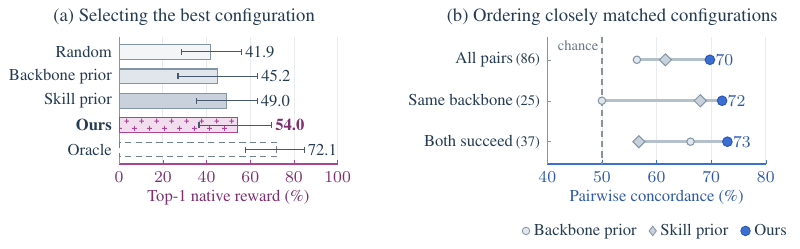}
\caption{Native execution on official SkillsBench runs. (a) Top-1 native reward with 95\% task-bootstrap intervals. (b) Within-task pairwise concordance on pairs with different rewards (pair counts in parentheses).}
\label{fig:e2e_native}
\end{figure}

\noindent\textbf{Results.} The critic's top choice reaches a native reward of 54.0\%, compared with 41.9\% for random selection, 45.2\% for the backbone prior, and 49.0\% for the skill prior, closing 40\% of the gap between random selection and the oracle (72.1\%, Figure~\ref{fig:e2e_native}a and Table~\ref{tab:e2e_native}). Within tasks, it orders 69.8\% of configuration pairs correctly, above both priors. The advantage holds for the hardest pairs: 72.0\% when both configurations share a backbone and differ only in skills, and 73.0\% when both succeed with a nonzero reward.

\subsection{AgentSelect}
\label{app:execution_baselines_agentselect}

\noindent\textbf{Setting.} The baselines are RAG-Qwen3.8-27B and Bundle-Ret. For ours, we use the top-ranked configuration of the full method (Stages~1--3). Each top-ranked configuration runs and is judged as in Appendix~\ref{app:e2e_runtime}. We also record turns and wall time per query, total tokens, and backbone API cost.

\begin{table}[h]
\centering
\caption{Execution on AgentSelect.}
\label{tab:execution_baselines_agentselect}
\small
\setlength{\tabcolsep}{3.4pt}
\arrayrulecolor{TableOneInk!65}
\begin{tabular}{@{}l*{10}{c}@{}}
\toprule
& \multicolumn{6}{c}{Execution quality ($\uparrow$)} & \multicolumn{4}{c}{Execution cost ($\downarrow$)} \\
\cmidrule(lr){2-7}\cmidrule(l){8-11}
Method & Total & Compl. & Corr. & Constr. & \makecell[b]{$\geq 8$\\(\%)} & Answers & Turns & \makecell[b]{Time\\(s)} & \makecell[b]{Tokens\\(K)} & \makecell[b]{Cost\\(CNY)} \\
\midrule
Bundle-Ret      & 7.10 & 2.58 & 2.86 & 1.66 & 48 & 22 & 3.96 & 106.2 & 257.6 & 0.62 \\
RAG-Qwen3.8-27B & 7.40 & 2.82 & 3.00 & 1.58 & 56 & 23 & \textbf{3.04} & 72.1 & 231.4 & 0.86 \\
Ours            & \textbf{8.72} & \textbf{3.32} & \textbf{3.52} & \textbf{1.88} & \textbf{72} & \textbf{25} & \textbf{3.04} & \textbf{42.2} & \textbf{180.6} & \textbf{0.50} \\
\bottomrule
\end{tabular}

\end{table}

\noindent\textbf{Results.} Table~\ref{tab:execution_baselines_agentselect} reports the judge's total score and its three parts, the share of queries scoring at least 8, and the number of queries that end with a final answer (Answers). Ours has the highest mean score (8.72, compared with 7.40 and 7.10) and the highest score in all three parts. 72\% of its queries score at least 8, compared with 56\% and 48\%. It uses 3.04 assistant turns per query, as many as RAG-Qwen3.8-27B and fewer than Bundle-Ret (3.96). Ours returns a final answer on all 25 queries, while RAG-Qwen3.8-27B and Bundle-Ret miss two and three. It also takes the least time (42 seconds per query, compared with 72 and 106) and uses 22\% to 30\% fewer tokens.

\begin{figure}[h]
\centering
\includegraphics[width=\linewidth]{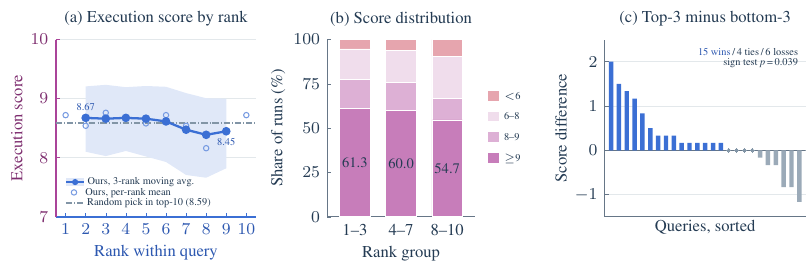}
\caption{Executing the top-10 list of ours on 25 AgentSelect queries (250 runs). (a) Execution score by rank: 3-rank moving average with a 95\% query-bootstrap band; rings are per-rank means; the dash-dotted line is the mean over all ten. (b) Share of runs per score band in each rank group. (c) Per-query mean of ranks 1--3 minus ranks 8--10, sorted.}
\label{fig:execution_top10}
\end{figure}

\noindent\textbf{Ranking within the top-10 list.} For each of the 25 queries, our full method returns a top-10 list, and we execute all ten configurations in the same harness (250 runs). The list is uniformly strong: the mean score is 8.59, every rank averages at least 8.16, and 73.6\% of runs score at least 8 (Figure~\ref{fig:execution_top10}). Quality concentrates at the top of the list. The 3-rank moving average falls from 8.67 (ranks 1--3) to 8.45 (ranks 8--10). From ranks 1--3 to 4--7 to 8--10, the share of runs scoring at least 8 falls from 77.3\% to 76.0\% to 66.7\%, and the share below 6 rises from 5.3\% to 6.0\% to 9.3\%. Per query, ranks 1--3 outscore ranks 8--10 in 15 queries and trail in 6 (4 ties; one-sided sign test, $p=0.039$).

\clearpage
\section{Case Study}
\label{app:case_study}

\begin{wrapfigure}{r}{0.3\textwidth}
  \vspace{-8pt}
  \centering
  \includegraphics[width=\linewidth]{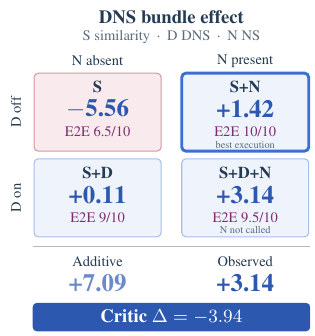}
  \vspace{-6mm}
  \caption{Bundle effects.}
  \label{fig:bundle_effect_case}
  \vspace{-2mm}
\end{wrapfigure}

We examine a query from the end-to-end evaluation to illustrate
how tool value depends on its companion resources. The user requests
a domain-similarity check, NS records, and a nameserver IP.
We fix DeepSeek-V4-Pro and the similarity tool $S$, then vary
DNS Lookup $D$ and NS Lookup $N$.
\noindent\textbf{Tool contributions depend on the bundle.}
The raw critic scores for $S$, $S{+}D$, $S{+}N$, and
$S{+}D{+}N$ are $-$5.56, 0.11, 1.42, and 3.14
(Figure~\ref{fig:bundle_effect_case}).
Adding $D$ to $S$ yields a 5.67 gain, but only 1.72 when
$N$ is already present, indicating diminishing returns from
overlapping lookup capabilities. An additive extrapolation would
predict 7.09 for the full bundle rather than the observed 3.14,
showing a clear non-additive interaction.
\noindent\textbf{Critic ranking agrees with execution quality.}
The end-to-end judge scores of $S$, $S{+}D$, $S{+}N$, and $S{+}D{+}N$ are
6.5, 9, 10, and 9.5. The critic ranks the two bundles with $N$
first and $S$ alone last, as execution does. The three-tool agent does not
call $N$, and $S{+}N$ achieves the best execution with fewer tools. This case
illustrates both context-dependent tool value and the agreement between the
critic's ranking and observed execution quality.

\end{document}